\documentclass[pmlr,twocolumn,10pt]{jmlr} 

\mlhtrack{proceedings}

\newif\iffinal
\finaltrue  

\iffinal
    \ifmlhneedspmlr
      \jmlrvolume{346}
      \jmlryear{2026}
    \fi
    \ifmlhfindings \jmlrproceedings{}{ML4H 2026 - Findings Track}\fi
    \ifmlhdemo     \jmlrproceedings{}{ML4H 2026 - Demo Track}\fi
    \jmlrworkshop{Machine Learning for Health (ML4H) 2026}
\else
    \jmlrproceedings{}{Submitted to ML4H 2026: \mlhtrackname}
    \jmlrworkshop{Machine Learning for Health (ML4H) 2026}
\fi

\usepackage{booktabs}
\usepackage{siunitx}

\usepackage[switch]{lineno}

\usepackage{dsfont}

\theorembodyfont{\upshape}
\theoremheaderfont{\scshape}
\theorempostheader{:}
\theoremsep{\newline}

\title[DynSurvX]{DynSurvX: Time-Resolved Explanations for \titlebreak Dynamic Survival Predictions}

\author{\Name{{Van Tuan} Nguyen} \Email{van-tuan.nguyen@inria.fr}\\
\addr Inria, Université Paris Cité, Inserm, HeKA, F-75015 Paris, France
\AND
\Name{Agathe Guilloux} \Email{agathe.guilloux@inria.fr}\\
\addr Inria, Université Paris Cité, Inserm, HeKA, F-75015 Paris, France}

\begin{document}

\maketitle

\ifmlhdemo\else
\begin{abstract}
Recent survival models integrating longitudinal data achieve strong predictive performance, but their predictions are often difficult to interpret, limiting their clinical use. We propose \texttt{DynSurvX}, a model-agnostic explainability framework for dynamic survival prediction with mixed static and longitudinal covariates. Based on Shapley values, \texttt{DynSurvX} introduces a hierarchical two-phase strategy that first identifies influential longitudinal features and then adaptively refines their temporal importance, enabling fine-grained explanations while reducing computational cost. The framework provides time-resolved explanations that evolve as new patient data become available. Experiments on simulated and two clinical datasets show varying agreement with theoretical importance across prediction models, with closer agreement for better-performing models, alongside clinically interpretable insights in real-world settings. It remains robust across prediction settings and reduces computation time by 41--80\% relative to the temporal explanation baseline, making dynamic explanations practical.
\end{abstract}


\paragraph*{Data and Code Availability} We used two real-world clinical datasets containing longitudinal patient data. The publicly available Mayo Clinic Primary Biliary Cirrhosis (PBCseq) dataset was accessed through the R package JMbayes~\citep{2017_JMbayes}. The MSK-Chord dataset is available through the \href{https://www.cbioportal.org/study/clinicalData?id=msk_chord_2024}{cBioPortal platform}, subject to data access approval. Our code is publicly available at \href{https://github.com/TeamHeKA/survinsights}{https://github.com/TeamHeKA/survinsights}.

\paragraph*{Institutional Review Board (IRB)} This research does not require IRB approval.

\section{Introduction}
\label{sec:intro}

Risk prediction, and more generally time-to-event prediction, plays a central role in clinical decision-making, informing treatment selection, patient stratification, and follow-up scheduling. In longitudinal care, these predictions are repeatedly updated as new biomarkers, imaging results, or treatment responses become available. Understanding not only how predictions evolve over time, but also which factors drive these changes, is therefore essential for trustworthy clinical decision-making.

Recent advances in survival analysis increasingly integrate longitudinal measurements, improving prognostic performance~\citep{lee2019dynamic,bleistein2024dynamic,mesinovic2026dysurv}. By leveraging repeated measurements, these models better capture disease progression and patient heterogeneity. However, this improved predictive performance often comes at the cost of interpretability: many approaches rely on latent representations or deep architectures, making them black boxes. Moreover, existing explainability methods are either tailored to specific model classes~\citep{lee2019dynamic} or restricted to survival models with static covariates~\citep{krzyzinski2023survshap,ter2024explainable,baniecki2025interpretable}, limiting their applicability in real-world clinical settings.

Most explainability methods rely on feature attribution to quantify the contribution of individual variables to model predictions. Among them, SHAP (SHapley Additive exPlanations)~\citep{lundberg2017unified}, based on cooperative game theory~\citep{shapley1953value}, is a widely used model-agnostic framework. Temporal extensions such as TimeSHAP~\citep{bento2021timeshap}, WindowSHAP~\citep{nayebi2023windowshap}, and OrdSHAP~\citep{hillordshap} explain both which features influence a prediction and when they do. However, these methods are primarily developed for regression and classification tasks. Moreover, temporal attribution is typically performed jointly over the feature--time space, incurring substantial computational cost that limits repeated explanations across successive prediction horizons.

\paragraph{Contributions and Paper Outline}

Our contributions are threefold:

\begin{itemize}

\item We introduce \texttt{DynSurvX}, a SHAP-based temporal attribution framework for survival models with longitudinal covariates. It explains the contributions of both static and longitudinal features while identifying when longitudinal observations influence survival predictions.

\item We formulate temporal explanation as a hierarchical attribution problem. The resulting two-phase strategy decouples feature screening from temporal refinement, substantially reducing the cost of temporal SHAP explanations while preserving explanation quality, making dynamic explanation practical.

\item Extensive experiments on simulated and real-world clinical datasets demonstrate that \texttt{DynSurvX} captures feature and temporal importance patterns with varying degrees of agreement with theoretical importance, while providing clinically interpretable explanations across different survival models.

\end{itemize}

The remainder of the paper is organized as follows. Section~\ref{sec:related_work} reviews related work. Section~\ref{sec:background} introduces the problem setting and technical preliminaries. Section~\ref{sec:DynSurvX} presents the proposed method. Section~\ref{sec:experiments} describes the experimental setup. Section~\ref{sec:results} presents the results, and Section~\ref{sec:conclusion} concludes the paper.

\section{Related Work}
\label{sec:related_work}

\paragraph{Survival models with longitudinal data}

Survival models increasingly incorporate longitudinal data to improve dynamic prediction. Classical joint models explicitly model the dependence between longitudinal trajectories and survival outcomes, whereas recent deep learning approaches learn complex temporal representations from repeated measurements~\citep{rustand2024fast,nguyen2024efficient,lee2019dynamic,moon2022survlatent,bleistein2024dynamic}. Although these models achieve strong predictive performance, their reliance on latent representations limits interpretability and prevents direct time-resolved attribution of clinically observed longitudinal variables.

\paragraph{Explainability in survival analysis}

Several explainability methods originally developed for regression and classification have been adapted to survival analysis by defining survival-specific prediction functions. Representative examples include SurvLIME~\citep{kovalev2020survlime} and SurvSHAP~\citep{krzyzinski2023survshap,ter2024explainable}. However, these methods are limited to survival models with static covariates and cannot explain how the importance of longitudinal measurements evolves throughout patient follow-up.

\paragraph{Explainability for temporal models}

Recent advances in explainable AI extend feature attribution to temporal models, explaining both which features influence a prediction and when they do. Representative methods include TimeSHAP~\citep{bento2021timeshap}, WindowSHAP~\citep{nayebi2023windowshap}, and OrdSHAP~\citep{hillordshap}. They perform attribution jointly over the feature--time space, requiring Shapley values to be estimated for feature--time combinations. To reduce the computational burden, TimeSHAP aggregates early observations, whereas WindowSHAP groups neighboring time steps into temporal windows. Nevertheless, temporal Shapley explanations remain computationally expensive for high-dimensional longitudinal data, making repeated explanations across prediction horizons impractical.

For longitudinal survival models, \citet{chen2025jointlime} introduce JointLIME, which uses a joint model as a local surrogate for black-box survival predictions. However, its explanations do not provide interval-specific attributions showing when longitudinal features influence the prediction.

\paragraph{Research gap}

Existing methods cannot jointly explain static and longitudinal features, identify when longitudinal information influences survival predictions, and efficiently scale temporal attribution. To address these limitations, we introduce \texttt{DynSurvX}, a model-agnostic Shapley framework that provides both overall and time-resolved explanations. Its hierarchical two-phase strategy substantially reduces the temporal attribution space while preserving the ability to localize informative temporal intervals.

\section{Defining Shapley Values for Dynamic Survival Prediction}
\label{sec:background}

This section introduces the dynamic survival prediction setting and the Shapley-value definitions underlying the proposed framework.

\paragraph{Dynamic survival prediction}
In longitudinal survival analysis, an individual's prognosis is updated as new measurements become available during follow-up. Consider an individual $i$ who is alive at prediction time $t_{\mathrm{P}}$. Up to this time, we observe $h$ static covariates $W^i=(W^i_1,\ldots,W^i_h)$
and $d$ longitudinal covariates measured at subject-specific times $t^i_1\leq\cdots\leq t^i_{m_i}\leq t_{\mathrm{P}}.
$ Their observed longitudinal history is denoted by
\[
X^i_{\leq t_{\mathrm{P}}}
=
\big(
X^i(t^i_1),\ldots,X^i(t^i_{m_i})
\big),
\qquad
X^i(t)\in\mathbb{R}^d.
\]

Given $W^i$ and $X^i_{\leq t_{\mathrm{P}}}$, a survival model $M$ (trained on censored observations) predicts quantities summarizing the conditional distribution of the remaining survival time $T^i-t_{\mathrm{P}}$. Common prediction targets include the conditional survival probability at a fixed horizon $t_H$,
\begin{equation*}
    \pi^i(t_H \mid t_{\mathrm{P}}; M)
=
\mathds{P}\!\left(
T^i \ge t_H
\,\middle|\,
X^i_{\le t_{\mathrm{P}}}, W^i, T^i \ge t_{\mathrm{P}};\, M
\right), 
\end{equation*}
and the expected remaining restricted survival time
\[\mathds{E}\!\left[T^i \wedge t_H - t_{\mathrm{P}} \,\middle|\, W^i, X^i_{\le t_{\mathrm{P}}}; M\right]=\int_{t_{\text{P}}}^{t_H} \pi^i(t \mid t_{\text{P}}; M)\, dt.\]
Alternative prediction functions considered in the literature include the cumulative hazard function~\citep{kovalev2020survlime}, the median survival time~\citep{ter2024explainable}, and the survival function itself~\citep{krzyzinski2023survshap,baniecki2025interpretable}. We therefore denote a generic scalar prediction by
\begin{equation}
\label{eq:pred_function}
f^M\!\left(W^i,X^i_{\leq t_{\mathrm{P}}}\right),
\end{equation}
which is the quantity to be explained.


\paragraph{Objective}
The objective of this work is to quantify the contribution of both static covariates $W^i$ and longitudinal measurements $X^i_{\le t_{\mathrm{P}}}$, including their temporal structure, to the prediction function
$f^M\!\left(W^i, X^i_{\le t_{\mathrm{P}}}\right)$ defined in~\eqref{eq:pred_function}. Since these explanations must be repeatedly updated in dynamic prediction settings, computational efficiency is essential.


\paragraph{Background of Shapley values} 
Originally introduced in cooperative game theory~\citep{shapley1953value}, Shapley values attribute the output of a function to its input features through their average marginal contributions over all possible feature coalitions. SHAP~\citep{lundberg2017unified} provides a model-agnostic framework for estimating these contributions for individual predictions. More recently, \citet{krzyzinski2023survshap,ter2024explainable} extended this idea to survival analysis by defining Shapley-based explanations for survival predictions, focusing primarily on static covariates.

We formulate Shapley values for the dynamic prediction function
$f^M(W^i,X^i_{\le t_{\mathrm{P}}})$ in~\eqref{eq:pred_function} along two levels: (i) overall feature importance, treating each longitudinal trajectory as a single feature, and (ii) temporal importance, identifying the contribution of specific time intervals within each trajectory.

\paragraph{Shapley values for overall importance} We first quantify the contribution of static covariates and the overall contribution of longitudinal features. Let $Y^i_{\leq t_{\mathrm{P}}}=(W^i,X^i_{\leq t_{\mathrm{P}}})$
denote the combined input at prediction time $t_{\mathrm{P}}$, consisting of $h$ static features and $d$ longitudinal features. Each longitudinal trajectory is treated as a single feature, yielding $p=h+d$ features in total. For any feature $\ell\in[\![1,p]\!]$, the corresponding Shapley value is defined by averaging its marginal contribution over all possible feature coalitions (i.e., subsets of features). Formally,
\begin{equation*}
\begin{aligned}
\label{eq:shap_overall_importance}
&\phi^M(\ell,Y^i_{\leq t_{\mathrm{P}}})\\&= \sum_{S \subset [\![1,p ]\!] \setminus \{\ell\}}
\kappa_p(S) 
\Big[
g^M_{Y^i_{\leq t_{\mathrm{P}}}}(S \cup \{\ell\}) - g^M_{Y^i_{\leq t_{\mathrm{P}}}}(S)
\Big]
\end{aligned}
\end{equation*}
where
\[
\kappa_p(S)=\frac{|S|!(p-|S|-1)!}{p!},
\]
and
\[
g^M_{Y^i_{\leq t_{\mathrm{P}}}}(S)
=
\mathds{E}
\!\left[
f^M(W,X_{\leq t_{\mathrm{P}}})
\,\middle|\,
(W,X_{\leq t_{\mathrm{P}}})_S
=
Y^i_{\leq t_{\mathrm{P}},S}
\right].
\]
Here, $g^M_{Y^i_{\leq t_{\mathrm{P}}}}(S)$ denotes the expected prediction with the features in coalition $S$ fixed to their observed values and the remaining features marginalized out. 



\paragraph{Shapley values for temporal importance}
While overall importance quantifies the contribution of an entire longitudinal trajectory, it does not indicate when the trajectory influences the prediction. We therefore partition each longitudinal trajectory into temporal intervals and assign Shapley values to feature--interval pairs, following the temporal attribution principle of WindowSHAP~\citep{nayebi2023windowshap}.

For a longitudinal feature $\ell\in\{1,\ldots,d\}$, let
\[
\Omega^\ell=\{\omega_1^\ell,\ldots,\omega_{T_\ell}^\ell\}
\]
be a partition of $[0,t_{\mathrm P}]$. The Shapley value associated with interval $\omega_k^\ell$ is defined as
\begin{align*}&\phi^M(\ell,\omega_k^\ell,Y^i_{\leq t_{\mathrm{P}}})
\\&=
\sum_{S^\ell \subset \Omega^\ell \setminus \{\omega_k^\ell\}}
\kappa(S^\ell, T^\ell)
\big[
g^M_{Y^i_{\leq t_{\mathrm{P}}}}(S^\ell \cup \{\omega_k^\ell\}) - g^M_{Y^i_{\leq t_{\mathrm{P}}}}(S^\ell)
\big],
\end{align*}
where
\[
\kappa(S^\ell,T_\ell)
=
\frac{|S^\ell|!(T_\ell-|S^\ell|-1)!}{T_\ell!},
\]
and
\[
g^M_{Y^i_{\le t_{\mathrm P}}}(S^\ell)
=
\mathds{E}
\!\left[
f^M(W,X_{\le t_{\mathrm P}})
\;\middle|\;
X_{\ell,\omega}=X^i_{\ell,\omega},
~\forall\,\omega\in S^\ell
\right].
\]
Here, $g^M_{Y^i_{\le t_{\mathrm P}}}(S^\ell)$ denotes the expected prediction with the intervals in $S^\ell$ fixed to their observed values and the remaining intervals marginalized out.

However, directly applying this formulation to all feature--interval pairs leads to a rapidly growing attribution space. The next section introduces a hierarchical two-phase strategy to address this challenge.

\section{DynSurvX}
\label{sec:DynSurvX}


To reduce the computational cost of temporal Shapley explanations, we formulate temporal explanation as a hierarchical attribution problem. \texttt{DynSurvX} first identifies the longitudinal features that warrant temporal analysis, and then refines only these selected features to localize when they influence the prediction. This two-phase strategy avoids unnecessary temporal attribution, making repeated explanations practical in dynamic prediction settings.

\paragraph{Phase 1: Feature screening}
The first stage identifies the longitudinal features that warrant temporal refinement based on their Shapley values for overall importance defined in Section~\ref{sec:background}. In this work, we instantiate this stage by adapting KernelSHAP~\citep{lundberg2017unified} to the mixed static--longitudinal setting, where each longitudinal trajectory is treated as a single feature represented by its observed trajectory up to the prediction time $t_{\mathrm P}$. To rank features on a common scale and facilitate thresholding, we define the normalized importance score

\[
I^M(\ell,Y^i_{\le t_{\mathrm P}})
=
\frac{|\phi^M(\ell,Y^i_{\le t_{\mathrm P}})|}
{\sum_{j=1}^{p}|\phi^M(j,Y^i_{\le t_{\mathrm P}})|}.
\]
Only longitudinal features whose importance exceeds a threshold $\delta_1$ are refined in Phase~2. To balance explanation fidelity and computational efficiency, $\delta_1$ is selected automatically using our proposed elbow method (Appendix~\ref{apd:optimaldelta}).

\paragraph{Phase 2: Adaptive temporal refinement} The second stage refines the temporal importance of the selected longitudinal features. In this work, we instantiate this stage by extending Dynamic WindowSHAP~\citep{nayebi2023windowshap} with an adaptive refinement strategy. Each selected trajectory is initialized as a single interval, which is then recursively refined. At each step, Shapley values are computed for the current feature--interval pairs and, as in Phase~1, normalized to obtain interval importance scores. Only intervals whose normalized importance exceeds a threshold $\delta_2$ are further subdivided. Unlike Dynamic WindowSHAP, \texttt{DynSurvX} updates $\delta_2$ adaptively during refinement, automatically determining an appropriate temporal resolution for each prediction. Refinement stops when no informative intervals remain or the maximum number of refinement steps is reached. The final explanation identifies both which longitudinal features influence the prediction and when these contributions occur. Algorithm~\ref{alg:DynSurvX} (Appendix~\ref{appendix:dynsurvx-algo}) summarizes the complete procedure.

\paragraph{Dynamic update of explanations}

In dynamic prediction, survival predictions are continuously updated as new longitudinal measurements become available. Meaningful explanations should therefore evolve accordingly, reflecting the changing contribution of patient information over time. Since these explanations must be recomputed repeatedly throughout follow-up, reducing the computational cost of temporal attribution is essential. Figure~\ref{fig:dyn_expl} (Appendix~\ref{appendix:dyn_expl}) illustrates how these explanations evolve across successive prediction times.



\paragraph{Local and global explanation}

\texttt{DynSurvX} produces local explanations that quantify patient-specific feature and temporal importance. For population-level analysis, these local explanations are aggregated over the explained individuals. For feature $\ell$, the global importance score at prediction time $t_{\mathrm P}$ is defined as
\[
\bar I^M(\ell,t_{\mathrm P})
=
\frac{1}{N_{\mathrm{exp}}}
\sum_{i=1}^{N_{\mathrm{exp}}}
I^M(\ell,Y^i_{\le t_{\mathrm P}}),
\]
where $N_{\mathrm{exp}}$ is the number of explained individuals. The same aggregation applies to temporal importance scores, together with 95\% variability intervals reflecting inter-individual variability.

\section{Experiments}
\label{sec:experiments}

\subsection{Experimental overview}


We conduct a comprehensive evaluation of \texttt{DynSurvX} from four complementary perspectives: (i) agreement with theoretical feature and temporal importance patterns in simulations, (ii) clinical interpretability in real-world settings, (iii) explanation behavior across prediction models, targets, and times, and (iv) comparison with existing methods in terms of explanation quality and computational efficiency. The evaluation spans one simulated and two clinical datasets, three prediction models, multiple prediction targets, and multiple prediction times. This section describes the datasets, prediction framework, and explanation baselines, while Section~\ref{sec:results} presents the results.


\subsection{Datasets}
We evaluate \texttt{DynSurvX} on one simulated dataset and two real-world longitudinal survival datasets. 


\paragraph{Simulation dataset.}

The simulated dataset is generated using an Ornstein--Uhlenbeck latent process influenced by static covariates and longitudinal trajectories~\citep{maller2009ornstein}. Static covariates are sampled from a multivariate normal distribution, while longitudinal trajectories are constructed by downsampling fractional Brownian motion. Survival outcomes are determined by the first hitting time of the latent process, with right censoring applied. This simulation setting allows us to derive theoretical temporal importance scores (Appendix~\ref{apd:true_importance_score}), which, although not directly equivalent to \texttt{DynSurvX} attributions, provide a reference for assessing the agreement in relative feature importance and temporal patterns.

\paragraph{PBCseq} The PBCseq dataset~\citep{dickson1989prognosis,murtaugh1994primary} contains 312 patients with primary biliary cirrhosis, a rare autoimmune liver disease. We consider five longitudinal features - serum bilirubin (serBilir), serum albumin (albumin), alkaline phosphatase (alkaline), aspartate aminotransferase (SGOT), and prothrombin time (protime) - and two static features, age and edema.

\paragraph{MSK-Chord}

The MSK-Chord dataset~\citep{jee2024automated} contains clinical data from patients treated at Memorial Sloan Kettering Cancer Center. We restrict our analysis to the breast cancer cohort, using two longitudinal tumour markers (CEA and CA15-3) and five static covariates: age, stage, HR status, HER2 status, and smoking history.


Complete dataset descriptions, including the summary table (Table~\ref{tab:data_stat}), are provided in Appendix~\ref{apd:dataset}.


\subsection{Prediction framework}

\paragraph{Prediction models.}
We evaluate \texttt{DynSurvX} using three state-of-the-art dynamic survival models: Dynamic DeepHit (DDH)~\citep{lee2019dynamic}, CoxSig~\citep{bleistein2024dynamic} and DySurv~\citep{mesinovic2026dysurv}. All three models are trained following their original implementations, with hyperparameters selected by validation. Implementation details are provided in Appendix~\ref{subappendix:pred_models}.

\paragraph{Prediction targets.}
Unlike existing survival explanation methods designed for a specific prediction target~\citep{kovalev2020survlime,ter2024explainable}, \texttt{DynSurvX} is agnostic to the choice of prediction target. Throughout the experiments, we consider two prediction targets: the expected remaining restricted survival time, used as the primary prediction target, and the median residual survival time~\citep{ter2024explainable}. This also enables us to assess the consistency of the resulting explanations across different prediction targets.

\paragraph{Predictive performance assessment.}

Since \texttt{DynSurvX} is a post-hoc explanation framework, meaningful interpretation requires sufficiently accurate prediction models. If the underlying survival model learns inaccurate associations between the inputs and prediction target, even a faithful explanation method can only reflect these associations, limiting the clinical trustworthiness of its explanations. We therefore assess predictive performance before interpreting the explanations.

Performance is evaluated using the time-dependent concordance index for discrimination (higher is better) and the Brier score for calibration (lower is better; Appendix~\ref{subappendix:metrics}) (see~\citet{bleistein2024dynamic}). Both metrics are evaluated at prediction times $t_{\mathrm P}$ selected as quantiles of the observed event-time distribution, over the interval $[t_{\mathrm P},\, t_{\mathrm P}+\delta t]$, where $\delta t$ ensures that at least one event occurs within the prediction horizon.

\subsection{Explanation baselines}
To benchmark our \texttt{DynSurvX} algorithm in both explanation quality and runtime, we compare it with JointLIME and WindowSHAP. For WindowSHAP, the survival prediction target is treated as a regression target, enabling the use of temporal Shapley-based explainers originally developed for regression tasks. Due to the large number of coalitions generated by WindowSHAP and resulting memory constraints, we restrict the analysis to a subset of coalitions. An additional experiment is conducted to determine an efficient number of evaluated coalitions (Figure~\ref{fig:optimalcoaltions}, Appendix~\ref{apd:optimalcoaltions}).

\section{Results}
\label{sec:results}
This section presents the experimental results following the evaluation protocol introduced in Section~\ref{sec:experiments}. We begin with the predictive performance of the underlying survival models, followed by the evaluation of \texttt{DynSurvX} in terms of agreement with theoretical importance, clinical relevance, robustness to different prediction targets, and comparison with existing explanation methods.

\subsection{Prediction model performance}

We first report the predictive performance of the underlying survival models, as this provides the context for interpreting the subsequent explanation results.


On the simulated dataset, CoxSig consistently outperforms DDH and DySurv in both discrimination and calibration across prediction times (Figure~\ref{fig:simu_perf}). On PBCseq, DDH and DySurv substantially outperform CoxSig, particularly in discrimination, with DySurv achieving slightly higher C-index values than DDH across most evaluation settings. On MSK-Chord, all three models exhibit comparable but moderate predictive performance, with DDH showing a slight advantage in discrimination (Figures~\ref{fig:PBC_perf} and~\ref{fig:MSK_perf} in Appendix~\ref{apd:additional_pred_perf}).

\begin{figure}[!h]
    \centering    
    \includegraphics[width=.97\linewidth]{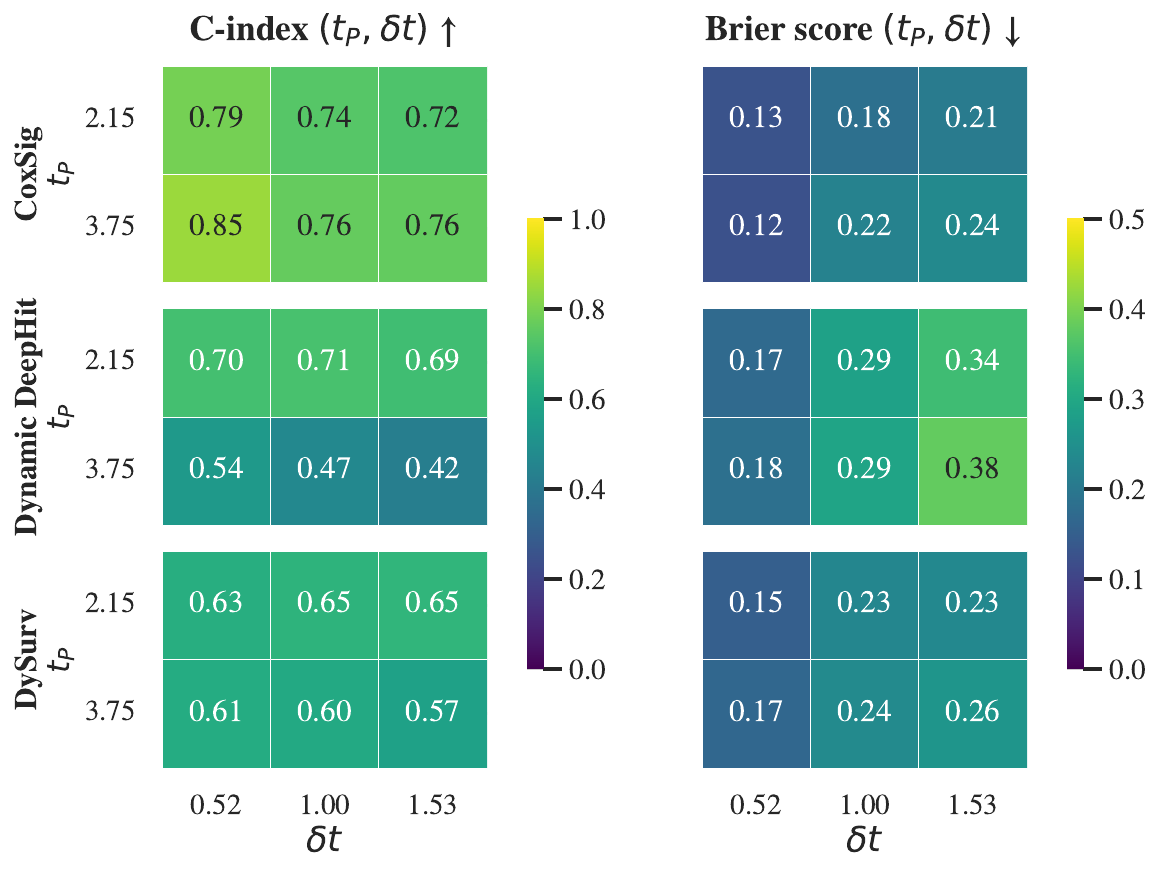}
    \caption{Simulated dataset -- Predictive performance. C-index (\textbf{left}) and Brier score (\textbf{right}) for CoxSig (\textbf{top}), DDH (\textbf{middle}), and DySurv (\textbf{bottom}) at prediction points $(t_{\mathrm P}, \delta t)$.} 
    \label{fig:simu_perf}
\end{figure}
\subsection{Agreement with theoretical importance}
In this section, we assess the agreement between \texttt{DynSurvX} explanations and theoretical importance patterns derived from the simulated data-generating process, examining both feature- and interval-wise importance across prediction models.

At the global level, \texttt{DynSurvX} identifies the main influential features and captures aspects of their temporal contributions, although the agreement with theoretical importance varies across prediction models (Figure~\ref{fig:simu_shap_value}). CoxSig shows the closest agreement with theoretical importance, identifying the dominant features and their early and recent temporal effects. In contrast, DDH identifies the most influential feature but places greater emphasis on recent observations, while DySurv distributes importance more broadly across features with less pronounced temporal variations. These differences are consistent with the models' predictive performance, with CoxSig achieving the highest C-index and lowest Brier score. Overall, these findings illustrate that \texttt{DynSurvX} provides model-dependent explanations, with the closest agreement with theoretical importance observed for the model exhibiting the strongest predictive performance. This highlights the importance of considering the predictive performance of the underlying models when interpreting explanations. Moreover, as additional longitudinal observations become available, \texttt{DynSurvX} captures changes in estimated feature importance across prediction times (Figures~\ref{fig:simu_shap_feat} and~\ref{fig:simu_shap_longi} in Appendix~\ref{apd:additional_expl_perf}).

\begin{figure*}
    \centering
    \includegraphics[width=\linewidth]{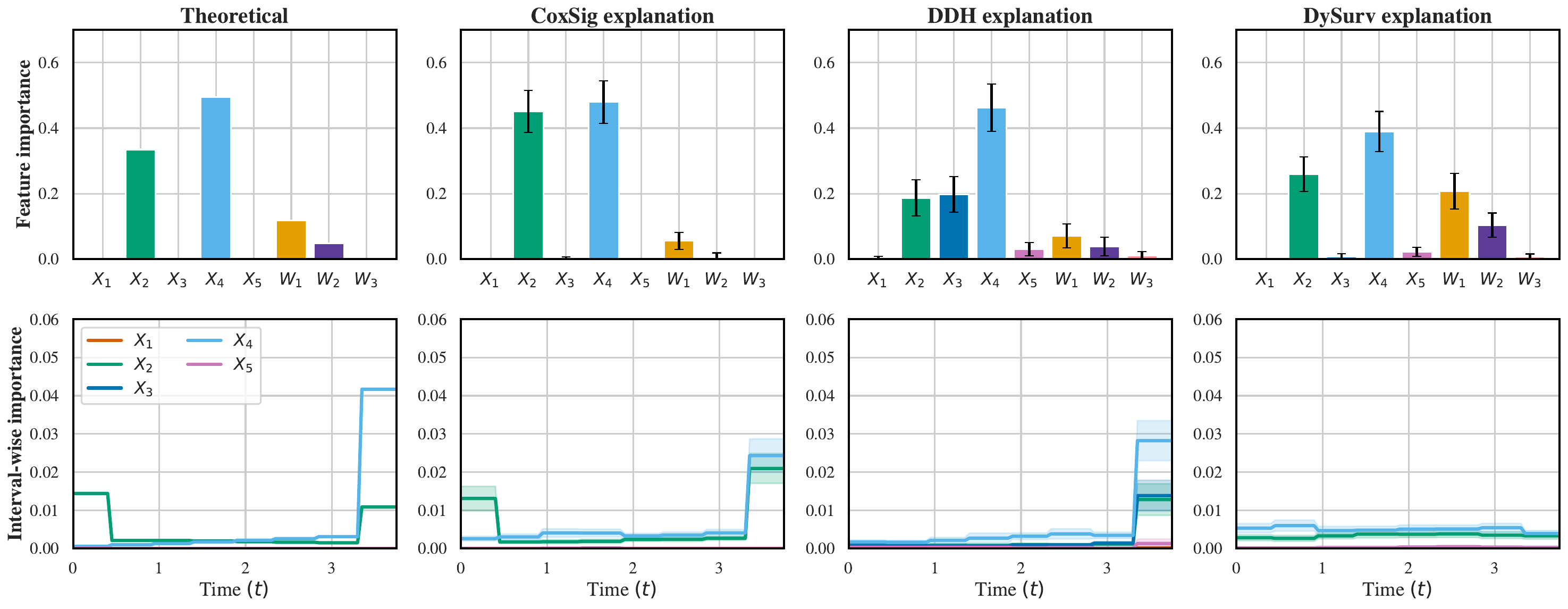}
    \caption{Simulated dataset -- \textbf{Feature importance} (\textbf{top}) and \textbf{interval-wise feature importance} (\textbf{bottom}). Theoretical values (\textbf{left}) and \texttt{DynSurvX} estimates for CoxSig, DDH, and DySurv (\textbf{three right panels}) at a prediction time.}
    \label{fig:simu_shap_value}
\end{figure*}

Finally, Figure~\ref{fig:simu_local} presents a local explanation for a representative individual. While the previous analyses assess agreement with theoretical importance at the population level, clinical decisions ultimately rely on individual explanations. The patient-specific temporal patterns identified by \texttt{DynSurvX} are consistent with the population-level explanations while providing individualized insights into the model's prediction.

\begin{figure}[!h]
    \centering
    \includegraphics[width=\linewidth]{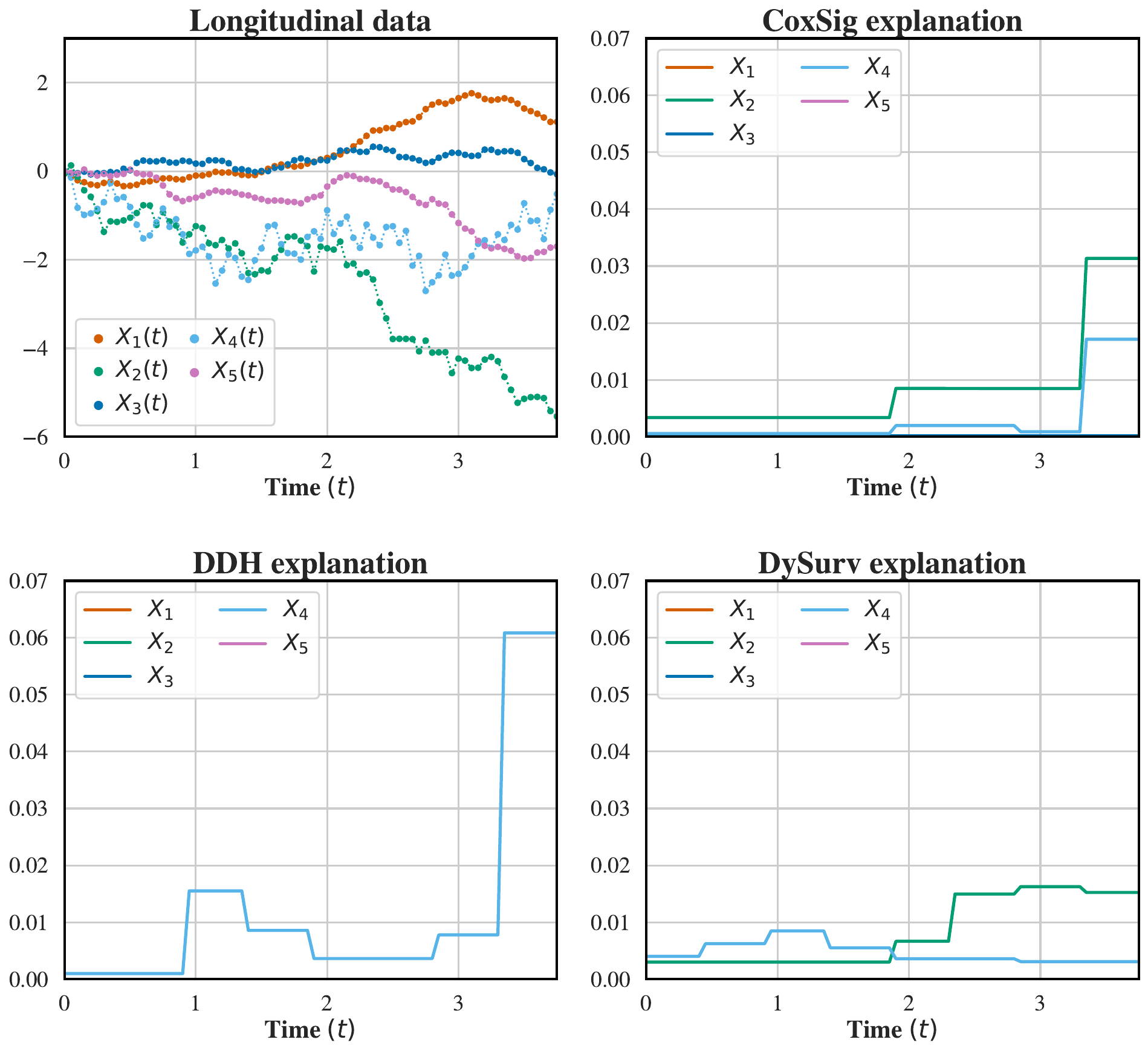}
    \caption{\textbf{Individual interval-wise feature importance}. Observed longitudinal data (\textbf{top, left}) and \texttt{DynSurvX} explanations for CoxSig (\textbf{top, right}), DDH (\textbf{bottom, left}), and DySurv (\textbf{bottom, right}).}
    \label{fig:simu_local}
\end{figure}


\subsection{Clinical interpretability}

We next assess whether the explanations produced by \texttt{DynSurvX} remain clinically meaningful when applied to real-world longitudinal survival datasets.

\paragraph{PBCseq} For DDH, \texttt{DynSurvX} identifies serum bilirubin as the dominant predictor, followed by serum albumin and prothrombin time (Figure~\ref{fig:PBC_shap}), consistent with established clinical knowledge~\citep{murtaugh1994primary}. Its temporal explanations reveal evolving feature importance, with recent measurements exerting greater influence on survival predictions. Similarly, DySurv highlights relevant longitudinal biomarkers, particularly prothrombin time and serum bilirubin, alongside static characteristics such as age and edema. In contrast, CoxSig emphasizes static covariates, particularly age, while relying less on longitudinal biomarkers, potentially reflecting difficulties in exploiting sparse longitudinal measurements. Overall, DDH and DySurv provide clinically consistent explanations, in line with their stronger predictive performance, while revealing distinct feature and temporal importance patterns.




\begin{figure}[h]
    \centering
    \includegraphics[width=\linewidth]{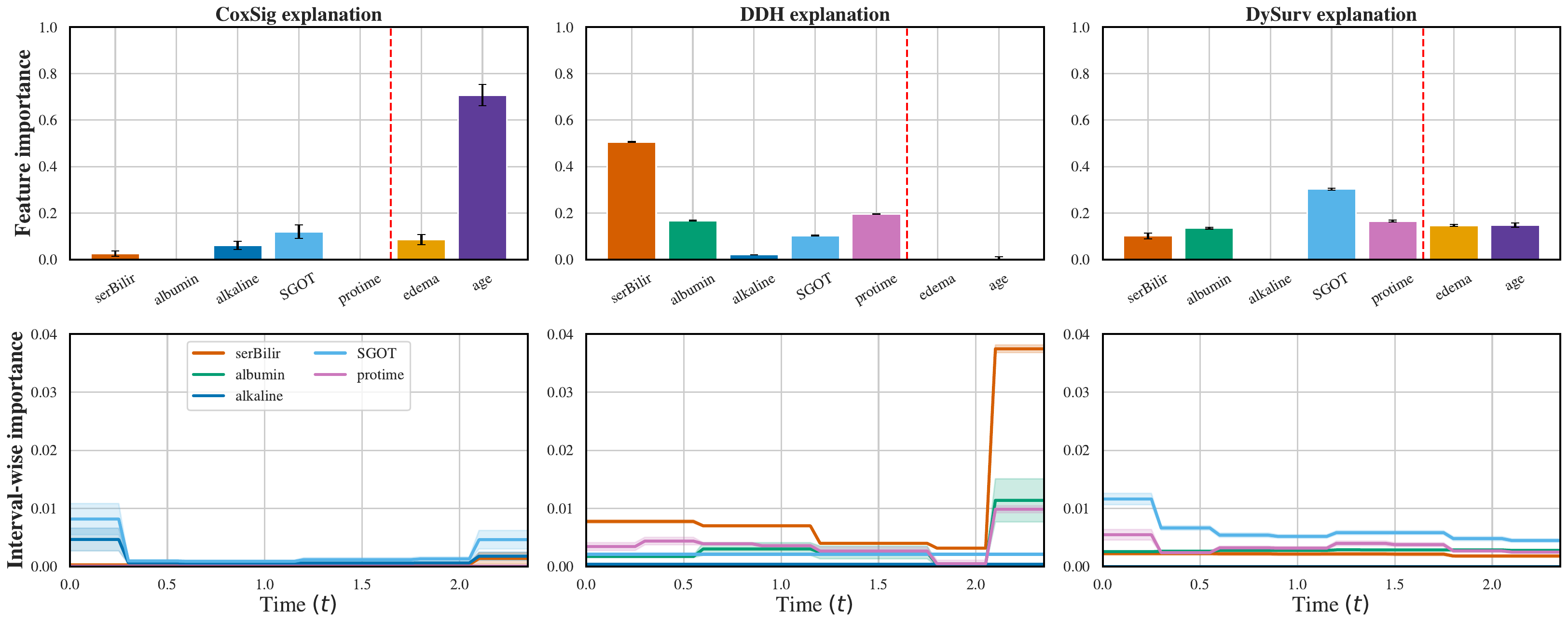}
    \caption{PBCseq dataset -- \textbf{Feature importance} (\textbf{top}) and \textbf{interval-wise feature importance} (\textbf{bottom}) estimated by \texttt{DynSurvX} for CoxSig (\textbf{left}), DDH (\textbf{middle}) and DySurv (\textbf{right}) at a prediction time. The red dashed line separates longitudinal and static features.}
    \label{fig:PBC_shap}
\end{figure}

\paragraph{MSK-Chord}

A similar analysis was conducted on the MSK-Chord dataset. CoxSig and DDH identify the longitudinal tumour markers CEA and CA15-3 as the most influential features (Figures~\ref{fig:MSK_shap_feat} and~\ref{fig:MSK_shap_longi} in Appendix~\ref{apd:additional_expl_perf}), consistent with their established prognostic relevance in breast cancer. While their global importance patterns are broadly similar, CoxSig also assigns moderate importance to patient age, whereas DDH relies predominantly on longitudinal biomarkers. In contrast, DySurv places greater emphasis on CEA, with a smaller contribution from CA15-3 and a broader distribution of importance across static covariates. Temporal explanations further reveal distinct patterns: CoxSig emphasizes early and recent measurements, DDH assigns greater importance to recent observations, and DySurv exhibits relatively stable contributions over time. Despite their comparable but moderate predictive performance, the three models exhibit substantial differences in their attribution patterns, highlighting the need for caution when interpreting these explanations.

Overall, the real-world experiments show that \texttt{DynSurvX} identifies clinically relevant feature and temporal importance patterns while revealing model-specific differences. The results further highlight the importance of considering predictive performance when interpreting explanations, as illustrated by the findings on PBC-Seq and the more heterogeneous attribution patterns observed on MSK-Chord, where predictive performance remains moderate.

\subsection{Robustness}
We further assess the stability of \texttt{DynSurvX} explanations across different design choices.

\paragraph{Choice of prediction target} To assess the sensitivity of the explanations to the prediction target, we consider the median residual survival time as an alternative to the expected remaining restricted survival time. When the median is well-defined, the resulting explanations exhibit broadly similar feature-level and temporal patterns across simulated and clinical datasets (Appendix~\ref{apd:additional_exp}). This demonstrates that the proposed framework is not tied to a specific scalar summary of the conditional survival distribution and can explain different prediction targets while producing consistent explanations.

\paragraph{Sensitivity analysis of threshold $\delta_1$}

The two-phase design of \texttt{DynSurvX} raises a potential concern: a longitudinal feature with low overall importance may nevertheless exhibit a strong localized temporal effect and therefore be discarded before temporal refinement. To assess both the robustness of the proposed feature-selection strategy and the sensitivity of the resulting explanations, we vary the threshold $\delta_1$ around the elbow-selected value.

On the PBCseq dataset using DDH, feature- and interval-wise importance patterns remain largely unchanged (Appendix~\ref{apd:sensitive_delta1}), indicating that the elbow-selected threshold provides a stable operating point while preserving clinically relevant temporal explanations. Nevertheless, this robustness does not eliminate the inherent limitation of the two-phase strategy. We therefore recommend treating $\delta_1$ as a tunable parameter and performing sensitivity analysis, particularly for datasets with sparse longitudinal signals.

\subsection{Comparison with existing methods}

Finally, we compare \texttt{DynSurvX} with existing explanation methods in terms of both explanation quality and computational efficiency.

\paragraph{Explanation quality} On the simulated dataset, \texttt{DynSurvX} more accurately recovers both overall and temporal importance than JointLIME and WindowSHAP (Figures~\ref{fig:simu_feat_compare} and~\ref{fig:simu_longi_compare}). JointLIME fails to identify several influential features, whereas Dynamic WindowSHAP detects the main longitudinal features but provides less accurate importance estimates and temporal localization. In contrast, \texttt{DynSurvX} closely matches the theoretical importance while jointly explaining both static and longitudinal features.

\begin{figure}[!h]
    \centering  \includegraphics[width=.9\linewidth]{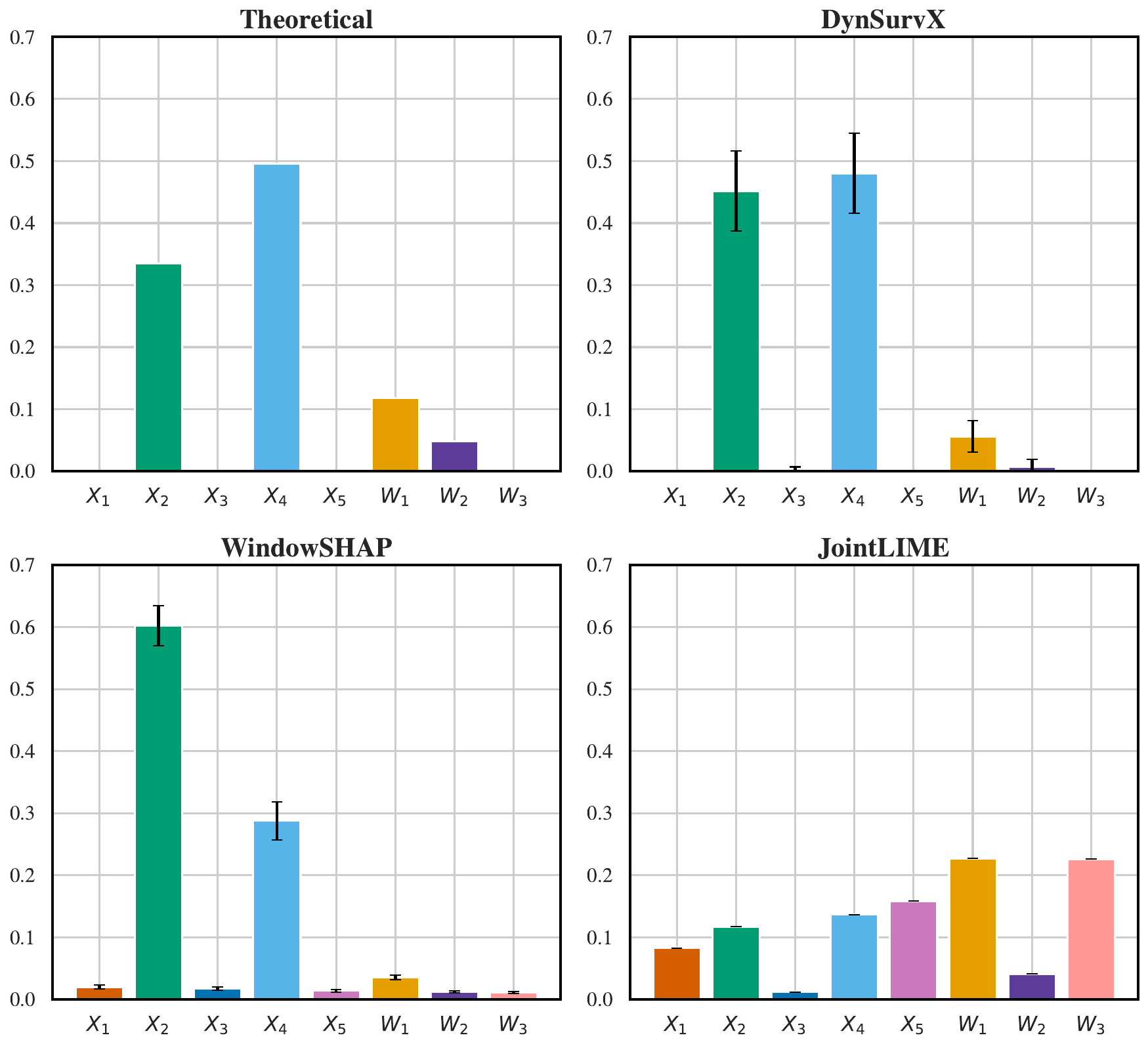}
    \caption{Simulated dataset -- \textbf{Feature importance} comparison. Theoretical values (\textbf{top}, \textbf{left}) and  estimates for \texttt{DynSurvX} (\textbf{top}, \textbf{right}), WindowSHAP (\textbf{bottom}, \textbf{left}) and JointLIME (\textbf{bottom}, \textbf{right}).}
    \label{fig:simu_feat_compare}
\end{figure}

\begin{figure}[!h]
    \centering  \includegraphics[width=\linewidth]{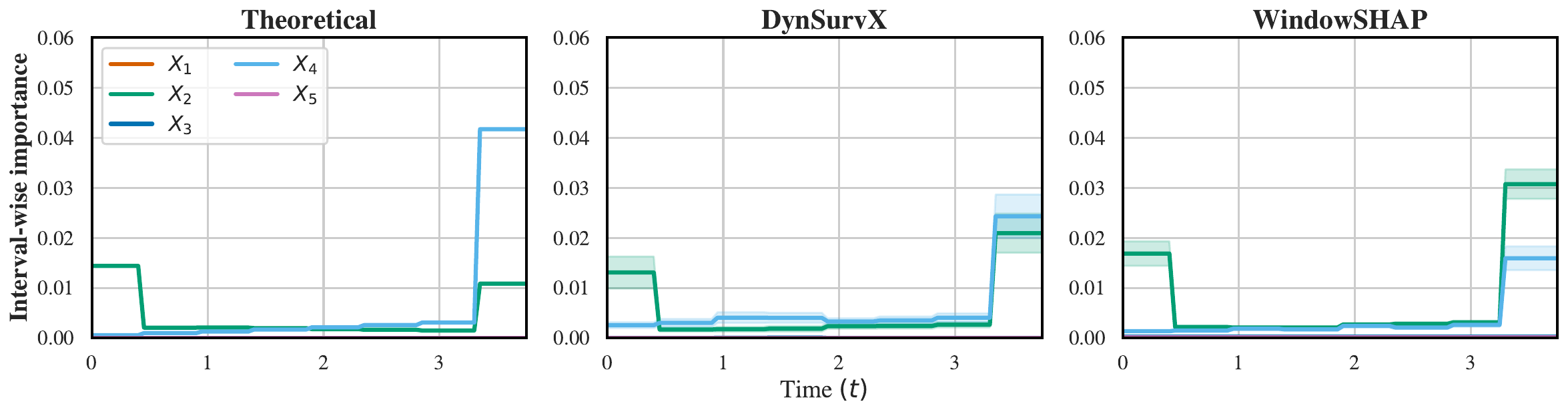}
    \caption{Simulated dataset - \textbf{Interval-wise feature importance}. Theoretical values (\textbf{left}) and estimates for \texttt{DynSurvX} (\textbf{middle}), WindowSHAP (\textbf{right}).}
    \label{fig:simu_longi_compare}
\end{figure}

\paragraph{Computational efficiency} Across the three datasets, \texttt{DynSurvX} requires 41--80\% less computation than WindowSHAP, particularly on datasets with many longitudinal features (Figure~\ref{fig:time_compare}). This demonstrates the efficiency of the hierarchical two-phase strategy, making repeated dynamic explanations practical.

\begin{figure}[!h]
    \centering  \includegraphics[width=.7\linewidth]{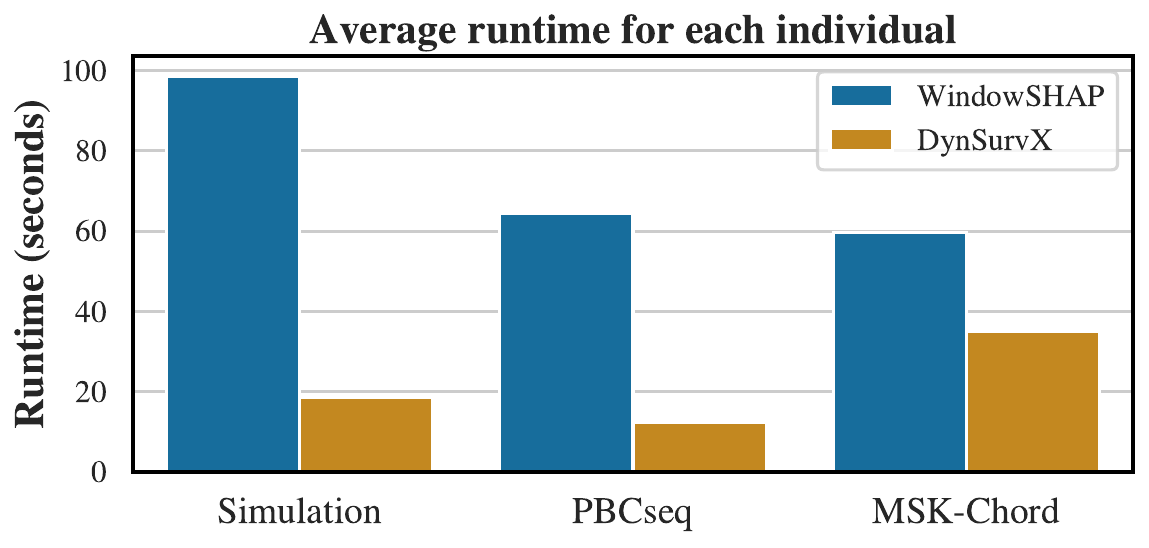}
    \caption{Runtime comparison.}
    \label{fig:time_compare}
\end{figure}

\section{Conclusion}
\label{sec:conclusion}
We introduced \texttt{DynSurvX}, a model-agnostic Shapley framework for explaining dynamic survival predictions with longitudinal features. Its hierarchical two-phase strategy combines feature screening with adaptive temporal refinement to provide overall and temporal explanations. Experiments demonstrate that, when applied to well-performing prediction models, \texttt{DynSurvX} achieves close agreement with theoretical importance patterns, provides clinically interpretable explanations, updates them coherently across prediction times, and remains robust across prediction settings. Compared with existing explanation methods, \texttt{DynSurvX} more accurately recovers theoretical importance and substantially reduces computational cost.

One limitation of \texttt{DynSurvX} is that interventional Shapley approximations may generate unrealistic perturbations when features are strongly dependent. Moreover, as a post-hoc framework, \texttt{DynSurvX} inherits limitations of the underlying prediction model, particularly under distribution shift. When predictive performance is poor, it remains unclear whether discrepancies in explanations arise from the prediction model, \texttt{DynSurvX} itself, or both. Future work will focus on improving explanation reliability through conditional Shapley estimators~\citep{aas2021explaining} and uncertainty-aware explanations~\citep{marx2023but}, as well as extending the framework to competing-risk and multi-state survival models~\citep{cottin2024ms} and evaluating additional prediction models.

\section*{Acknowledgments \& Funding}
For this work, VTN and AG have benefited from support from the National Agency for Research under the France 2030 program, reference ANR-22-PESN-0016. AG received funding from the MEDITWIN Bpifrance i-Demo project as part of the France 2030 program.

\newpage
\bibliography{ref}

\onecolumn
\appendix

\section{DynSurvX algorithm}
\label{appendix:dynsurvx}
\subsection{Algorithm}
\label{appendix:dynsurvx-algo}
\begin{algorithm}[]
\footnotesize
\caption{DynSurvX}\label{alg:DynSurvX}
\vspace{0.1in}
\hrule
\vspace{0.1in} 
\KwIn{Model $M$, prediction time $t_{\text{P}}$, observation of individual $i$: $Y^i_{\le t_P} = (W^i, X^i_{\le t_P})$ including static feature $W^i$ and
longitudinal feature $X^i_{\le t_P} = \big(
X^i(t^i_1), \ldots, X^i(t^i_{m_i})
\big)$, importance threshold control $\xi$, maximum number of split $n_s^{\text{max}}$}
\vspace{0.1in}
\textbf{Phase 1: Feature selection}

\vspace{0.04in}
$\{\phi^M(\ell,Y^i_{\le t_P})\}_{\ell=1}^p \leftarrow \texttt{KernelSHAP}(f^M, Y^i_{\le t_P})$

\vspace{0.03in}
\For{$\ell \in \{1,\ldots,p\}$}{
$ I^M(\ell,Y^i_{\le t_{\mathrm{P}}})
\leftarrow
\dfrac{|\phi^M(\ell,Y^i_{\le t_{\mathrm{P}}})|}
{\sum_{j=1}^p |\phi^M(j,Y^i_{\le t_{\mathrm{P}}})|}$\;
}

\vspace{0.04in}
\texttt{Select importance threshold $\delta_1$ via the elbow method}

\vspace{0.04in}
$\Psi \leftarrow \{\ell \in \{h+1,\ldots,p\}: I^M(\ell,Y^i_{\le t_{\mathrm{P}}}) > \delta_1\}$\;

\vspace{0.1in}
\textbf{Phase 2: Adaptive time window selection} \emph{(see Figure~\ref{fig:demo_phase2})}

\vspace{0.04in}
\For{$\ell \in \Psi$}{

\vspace{0.04in}
\texttt{Initialize boundary indices and threshold} 

\vspace{0.02in}
$B^\ell \leftarrow \{1, m_i\}$; \, $T^\ell = |B^\ell| - 1$; $\delta_2 = \min(\xi, \frac{1}{T^\ell})$

\vspace{0.05in}
\For{$q \in \{1, \ldots, n_s^{\text{max}}\}$}{

    \vspace{0.04in}
    \texttt{Build time windows from adjacent boundaries:}
    
    \vspace{0.02in}
    \For{$k \in \{1, \ldots, T^\ell\}$}{
    $\omega_k^\ell \leftarrow \big\{ t^i_\nu : B_k^\ell \leq \nu \leq B_{k+1}^\ell\big\}$
    \vspace{0.03in}
    }

    \vspace{0.03in}
    $\Omega^\ell = \big\{ \omega_k^\ell : 1 \leq k \leq T_\ell$ \big\}
    
    \vspace{0.05in}
    \texttt{Compute window Shapley values:}
    
    \vspace{0.03in} 
    {\footnotesize
    $\big\{\phi^M(\ell,\omega_k^\ell,Y^i_{\leq t_{\mathrm{P}}})\big\}_{k=1}^{T^\ell} \leftarrow \texttt{WindowSHAP}\Big(f^M, Y^i_{\leq t_{\mathrm{P}}}, \Omega^\ell\Big)$}

    \vspace{0.03in}
    \For{$k \in \{1, \ldots, T^\ell\}$}{
    $I^M(\ell, \omega_k^\ell, Y^i_{\leq t_{\mathrm{P}}}) = \frac{\big|\phi^M(\ell,\omega_k^\ell,Y^i_{\leq t_{\mathrm{P}}})\big|}{\sum_{j=1}^{|B^\ell| - 1} \big|\phi^M(\ell,\omega_j^\ell,Y^i_{\leq t_{\mathrm{P}}})\big|} $ }

    \vspace{0.03in}
    \texttt{Window spliting:}

    \vspace{0.03in}
    $B^\ell_{\text{new}} \leftarrow B^\ell$\;

    \vspace{0.03in}
    \For{$k \in \{1, \ldots, T^\ell\}$}{

    \vspace{0.03in}
    $\Delta B^\ell_k = B^\ell_{k+1} - B^\ell_k$
    
    \vspace{0.03in}
    \If{$I^M(\ell,\omega^\ell_k,Y^i_{\le t_{\mathrm{P}}}) \ge \delta_2$}{

    \vspace{0.05in}
    $b_{\text{mid}} \leftarrow \left\lfloor \dfrac{B^\ell_{k + 1}+B^\ell_k}{2}\right\rfloor$\; 
    
    Add $b_{\text{mid}}$ to $B^\ell_{\text{new}}$\;
    }
    }

    \vspace{0.03in}
    \If{$B^\ell_{\text{new}} = B^\ell$}{
    \vspace{0.03in}
    \textbf{break}\; \emph{(No further split.)}
    \vspace{0.03in}
    }
    \vspace{0.03in}
    $B^\ell \leftarrow \texttt{sort}(B^\ell_{\text{new}})$; \, $T^\ell = |B^\ell| - 1$; \, $\delta_2 = \min(\xi, \frac{1}{T^\ell})$
    }
}

\vspace{0.03in}
\Return{$\Psi,\ \{\Omega^\ell\}_{\ell\in\Psi},\ \{I^M(\ell,Y^i_{\le t_{\mathrm{P}}})\}_{\ell=1}^p,\ \{I^M(\ell,\omega,Y^i_{\le t_{\mathrm{P}}})\}$}\

\vspace{0.03in}
\hrule

\end{algorithm}

\begin{figure}[!ht]
    \centering
    \includegraphics[width=0.45\textwidth]{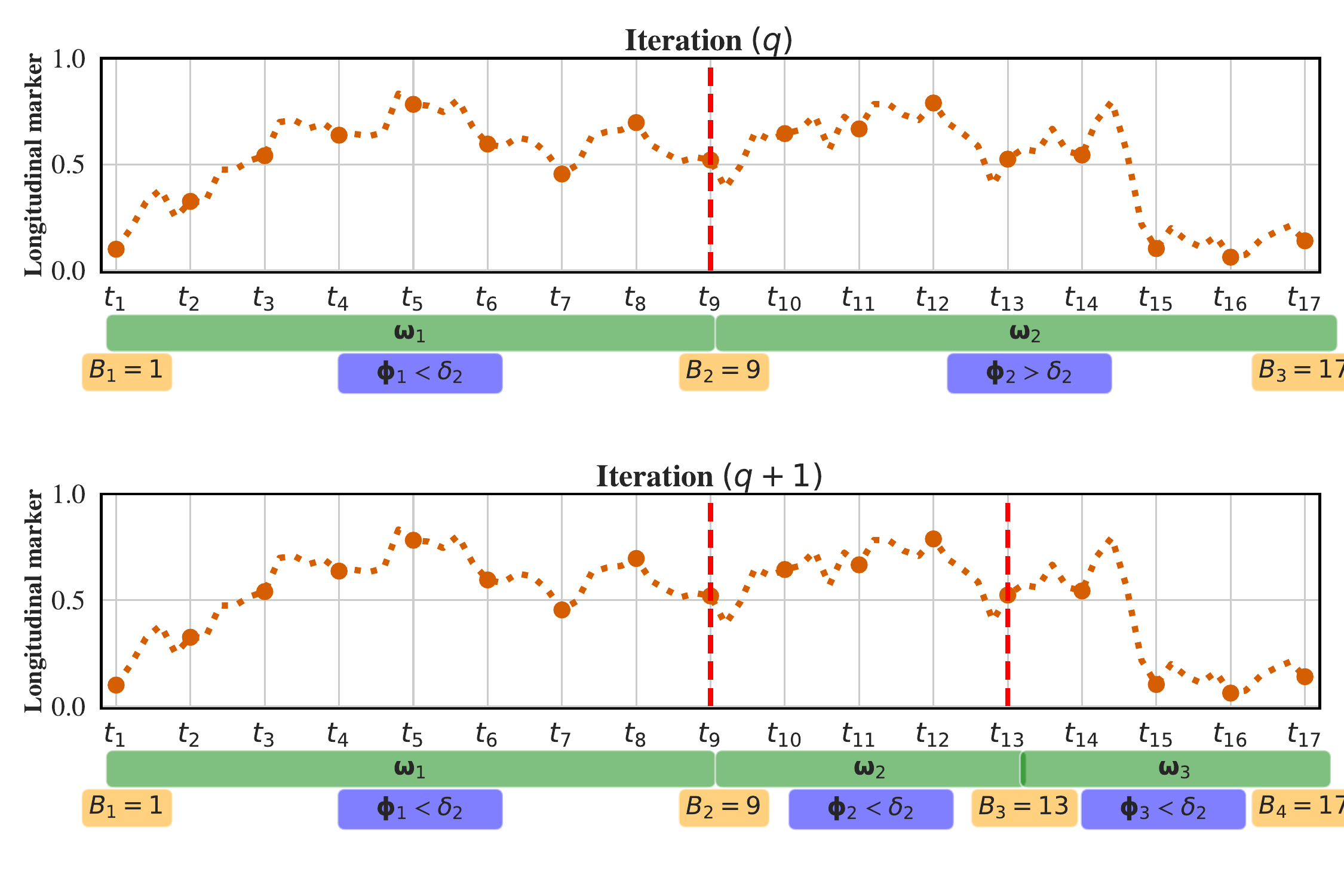}
    \caption{Illustration of the \texttt{DynSurvX} algorithm during Phase~2 on a longitudinal trajectory (in orange), between two consecutive splitting iterations $(q)$ and $(q+1)$. At iteration $(q)$, the second time window $\omega_2$ has an importance score exceeding the threshold $\delta_2$ and is therefore split into two smaller windows, while the first time window $\omega_1$ remains unchanged. The algorithm terminates at iteration $(q+1)$, as all window importance scores fall below the threshold $\delta_2$.
}
    \label{fig:demo_phase2}
\end{figure}

\subsection{Dynamic update of explanations}
\label{appendix:dyn_expl}
\begin{figure}[!ht]
    \centering
    \includegraphics[width=0.5\textwidth]{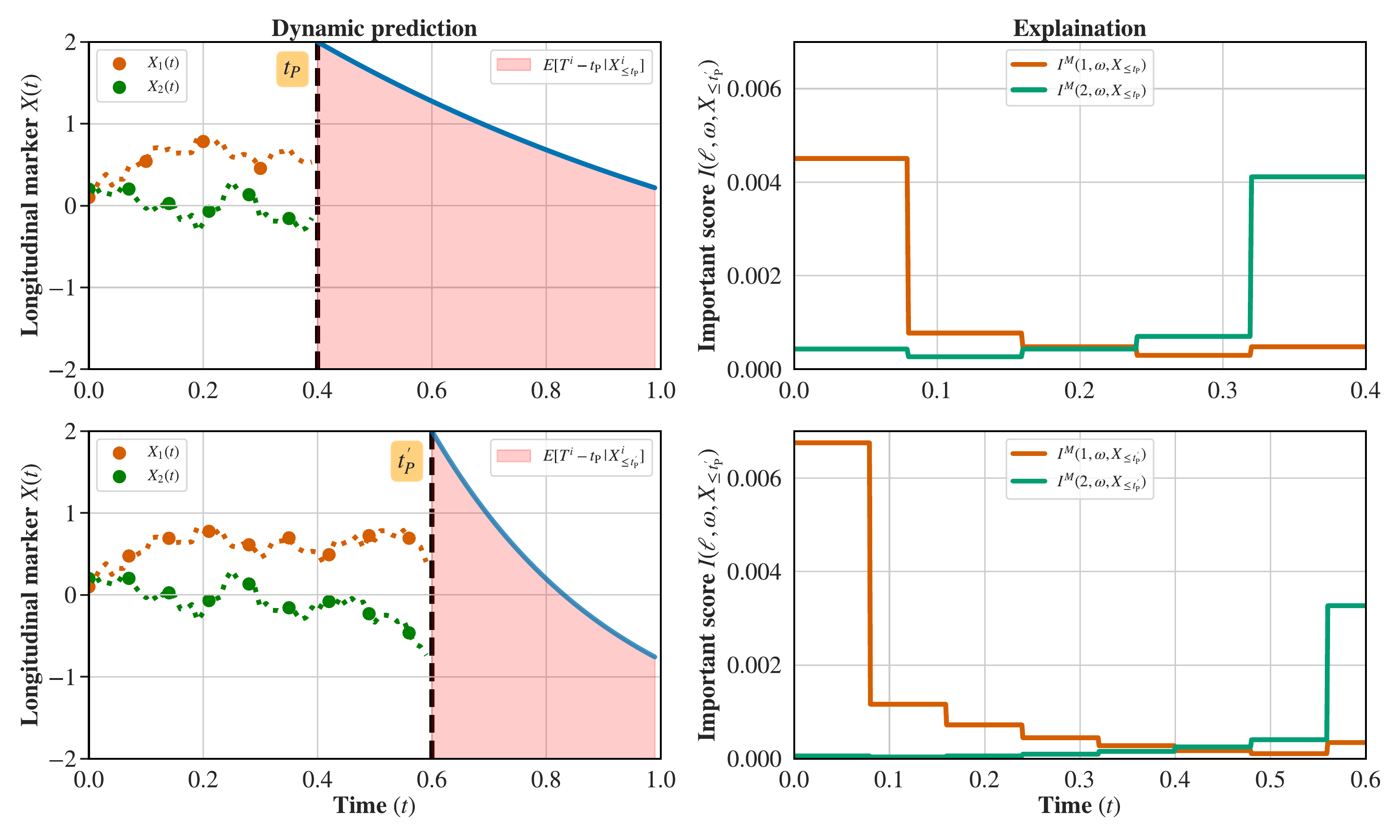}
    \caption{Dynamic explanations at two prediction times: $t_{\mathrm{P}}$ (\textbf{top}) and $t'_{\mathrm{P}}$ (\textbf{bottom}).
    Left panels show the observed longitudinal trajectories up to the prediction time and the corresponding expected remaining restricted survival time. Right panels display normalized interval-specific Shapley values produced by \texttt{DynSurvX}, highlighting how feature–time contributions evolve as new observations become available.}
    \label{fig:dyn_expl}
\end{figure}

\section{Details of Experiments and Datasets}
\label{apd:dataset}
\subsection{Simulation setting}
\label{apd:simu}
We describe in detail the simulation design used in our experiments. We fix
$h=3$ static covariates and $d=5$ longitudinal features. Individuals are assumed to be independent and identically distributed. In the following, we describe
the data-generating process for a single individual and drop the index
$i$ for notational simplicity.

\paragraph{Static covariates.} The static covariates $W \in \mathbb{R}^h$ are generated from a multivariate normal distribution with Toeplitz covariance:
\[
W \sim \mathcal{N}(0, \Sigma), \qquad
\Sigma_{jk} = \rho^{|j-k|}, \quad \rho= 0.5, \quad j,k = 1,\dots,h.
\] 

\paragraph{Longitudinal process.} The $d$-dimensional process $X(t) = \big(X^1(t),\ldots,X^d(t)\big)$ -- corresponding to the longitudinal features -- is obtained by downsampling a realization of a fractional Brownian motion with Hurst parameters

\[H = (0.9, 0.45, 0.85, 0.35, 0.8).\]
Observations are recorded on a regular time grid
$\{t_1 = 0, t_2 = \Delta\tau, \ldots , t_n = (n-1) \Delta\tau, \ldots\}$ with step size $\Delta\tau = 0.05$.

\paragraph{Latent risk process.}
The latent risk process $w$ is defined as
\begin{equation*}
    \label{eq:experiment_1_cde}
    w(t) = \sum\limits_{\ell=1}^d w^\ell(t)
\end{equation*} 
and each component $w^\ell$ (for $\ell = 1, \dots, d$) evolves according to an Ornstein--Uhlenbeck stochastic differential equation
\begin{equation*}
    \Delta w^\ell(t) = - \theta^\ell [w^\ell(t)-\mu^\ell]\Delta t +  \alpha^\ell \Delta X^\ell(t) + \sigma \Delta B^\ell(t) + \gamma W,
\end{equation*}
with initial condition $w^\ell(0)=1$. Here,  $(B^1, \ldots, B^d)$ denotes a vector of correlated Brownian motions. The model parameters are specified as
\[
d = 5, \qquad
\theta = (0, -0.3, 0, 0.2, 0), \qquad
\mu = 0.1, \qquad
\alpha = (0, 1, 0, 5, 0), \qquad
\sigma = 1
\]
and 
\[
\gamma^j \sim \mathcal{U}[-5, 5], \quad j = 1, \dots, h.
\]

\paragraph{Event time and censoring.}
The survival time is defined as the first hitting time of a threshold $w_\star = 12$, subject to censoring at time $t_C = 10$:
\[T = \min \{ 0 < t < t_C  \,|\, w(t) \geq w_\star\}.\]

If the threshold is not reached before $t_C$, the observation is right-censored at $t_C$. Conditional on the observed event or censoring time $T$, the longitudinal process is observed on the grid $t_1 \le \cdots \le t_n \le T$. 

\paragraph{Supplementary Figures.}
Figure~\ref{fig:path_rnd_individual} illustrates sample paths of the longitudinal process and the corresponding latent risk trajectory for a randomly selected individual.

\begin{figure}[h!]
    \centering
\includegraphics[width=0.5\textwidth]{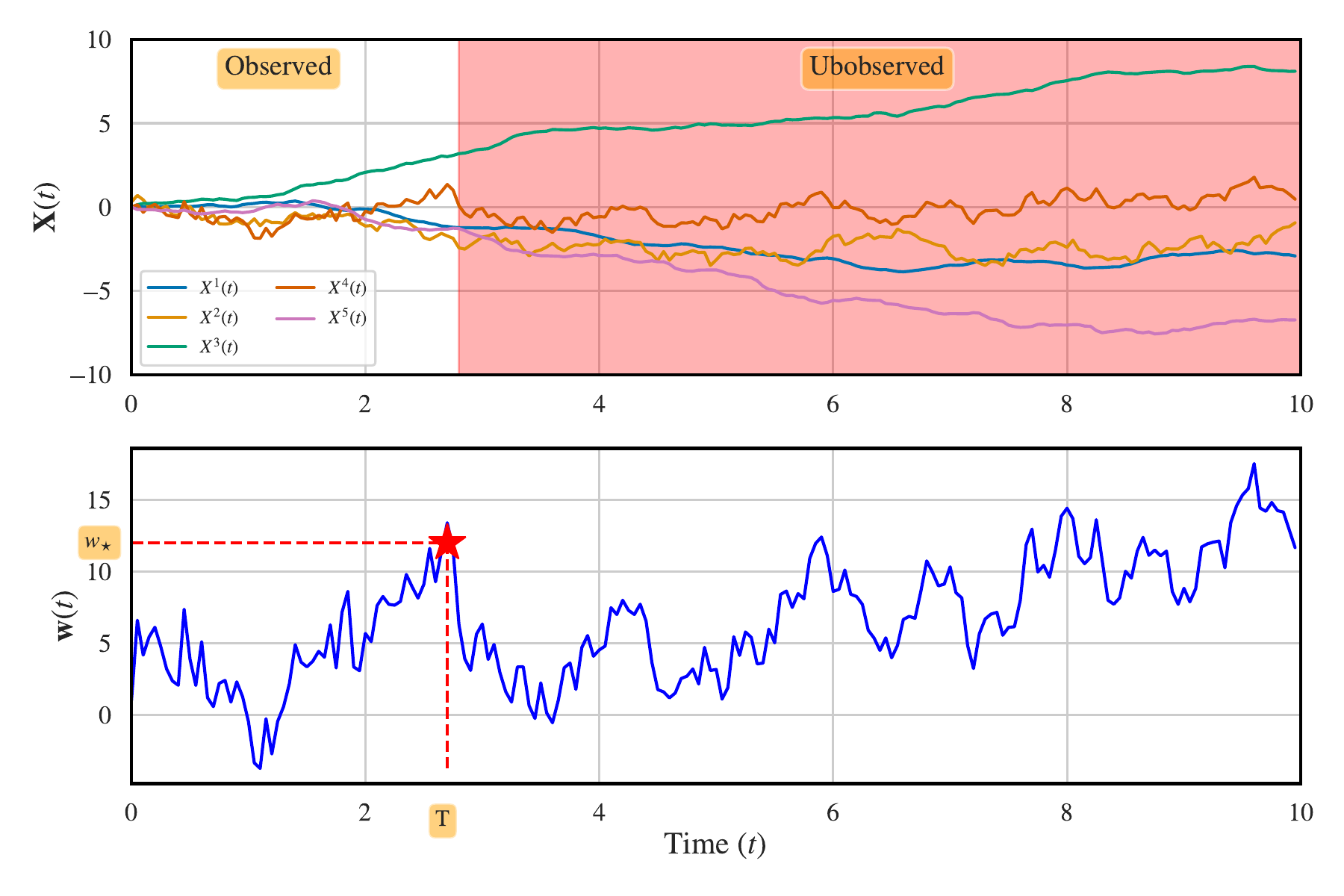}
\caption{Top: longitudinal trajectories $X(t)$ for a randomly selected individual. Bottom: latent risk process $w(t)$. The red star indicates the first hitting time of the threshold value $w_\star = 12$.}
\label{fig:path_rnd_individual}
\end{figure} 
Figure~\ref{fig:simu_data} illustrates the simulated trajectories and the distribution of event times.
\begin{figure}[!h]
\centering
\includegraphics[width=1.\linewidth]{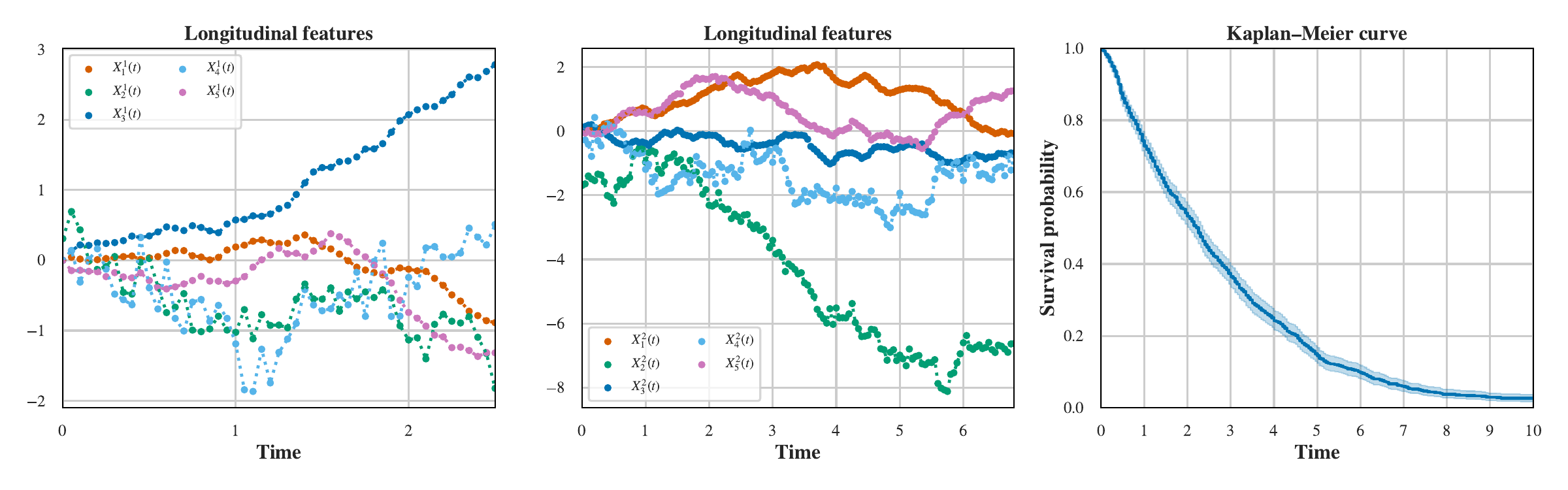}
\caption{Simulated dataset -- Example longitudinal trajectories for two individuals, recorded up to their observed times (\textbf{left}, \textbf{middle}) and the Kaplan–Meier estimate of the event-time distribution (\textbf{right}).}
\label{fig:simu_data}
\end{figure}
\subsection{Computation of the relative contributions}
\label{apd:true_importance_score}
On a regular time grid $\{t_1=0,  t_2=\Delta\tau, \ldots,  t_n = (n - 1)\Delta\tau\}$ with step size $\Delta\tau$, the discrete-time OU recursion can be expressed as
\begin{align*}
w^\ell(t_\kappa)
= \sum_{\iota=1}^\kappa &
  (-1)^{\mathds{1}_{\kappa=\iota}}
  \alpha^\ell
  \big[\theta^\ell \Delta\tau (1 - \theta^\ell \Delta\tau)^{\kappa-\iota-1}\big]^{\mathds{1}_{\iota \neq \kappa}}
  X^\ell(t_\iota) + M^\ell(t_\iota) + \gamma W,
\end{align*}
where
\[
M^\ell(t_\iota)
= (1 - \theta^\ell \Delta\tau)^{\kappa - \iota}
  \big[\theta^\ell \mu^\ell \Delta\tau + \sigma \Delta B^\ell(t_\iota)\big].
\]

The relative contribution of $X^\ell(t_\iota)$ to $w(t^\kappa)$ can then be quantified by the importance score
\begin{align*}
    I\!\left(X_\ell(t_\iota), w(t_\kappa)\right) = \frac{\big | \alpha^\ell \big[\theta^\ell \Delta\tau (1 - \theta^\ell \Delta\tau)^{\kappa-\iota-1}\big]^{\mathds{1}_{\kappa \neq \iota} } \big |}{\sum_{v=1}^d \sum_{u=1}^\kappa \big | \alpha^v \big[\theta^v \Delta\tau (1 - \theta^v \Delta\tau)^{\kappa-u-1}\big]^{\mathds{1}_{u \neq \kappa}} \big | + \big |\sum_{v=1}^d\sum_{u=1}^\kappa M^v(t_u)\big|+ \sum_{p=1}^h |\gamma^p|}.
\end{align*}
The average importance of $X^\ell(t_\iota)$ over all subsequent time points up to $t_\zeta > t_\iota$ is given by
\begin{align*}
    I\!\left(X_\ell(t_\iota)\right) = \frac{1}{\zeta - \iota} \sum_{\kappa = \iota + 1}^\zeta I\!\left(X_\ell(t_\iota), w(t_\kappa)\right).
\end{align*}

\subsection{PBCseq}
\label{apd:PBC_data}
The PBCseq dataset~\citep{dickson1989prognosis,murtaugh1994primary} follows 312 patients diagnosed with primary biliary cirrhosis, a rare autoimmune liver disease, and is widely used as a benchmark for longitudinal survival analysis.

\paragraph{Static covariates.} Several baseline covariates (e.g., sex, treatment assignment, histological stage) are recorded once at study entry; in our experiments, we retain age and edema status as static features and exclude the remaining baseline variables for simplicity.

\paragraph{Longitudinal data.} Clinical biomarkers are measured repeatedly during follow-up. In our experiments, we consider the following five longitudinal biomarkers:
\begin{itemize}
    \item serBilir: serum bilirubin in mg/dL,
    \item albumin: serum albumin in mg/dL,
    \item alkaline: alkaline phosphatase in U/L,
    \item SGOT: aspartate aminotransferase in U/mL,
    \item protime: prothrombin time in seconds.
\end{itemize}
The average number of measurement time points per patient is approximately 6.52. Selected biomarkers are log-transformed prior to model training to reduce skewness and improve numerical stability.
\paragraph{Event definition.}
The event time is defined as the number of days from study registration to the earliest occurrence of death or liver transplantation.

\paragraph{Censoring.}
Patients who are alive and transplant-free at the end of follow-up are treated as right-censored. The dataset exhibits a high censoring rate of approximately $57\%$, which is typical for long-term observational studies in survival analysis.

\paragraph{Supplementary figures.}
Additional illustrations are provided in Figure~\ref{fig:PBC_appendix_path_hist}, including example longitudinal trajectories up to the observed time for two selected patients and the empirical distribution of event times across the cohort.

\begin{figure}[!h]
\centering
\includegraphics[width=1.\linewidth]{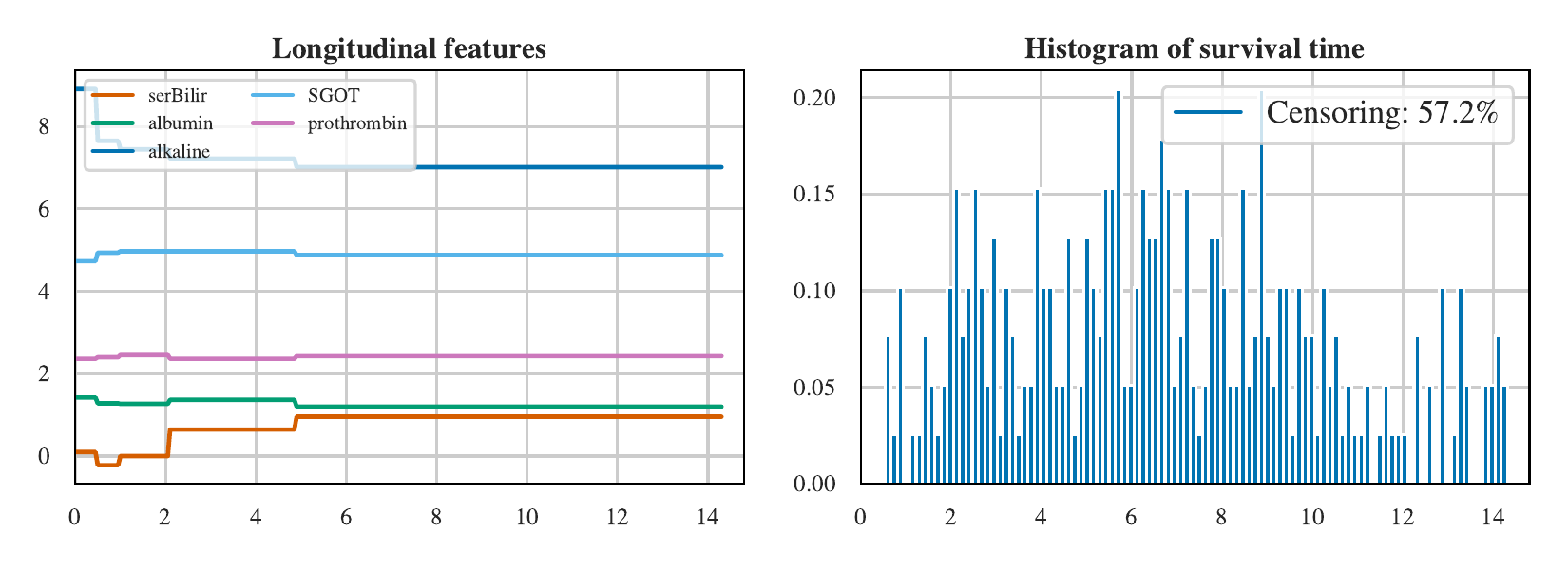}
\caption{PBCseq dataset -- Example longitudinal trajectories for two individuals, recorded up to their observed times (\textbf{left}, \textbf{middle}) and the Kaplan–Meier estimate of the event-time distribution (\textbf{right}).}
\label{fig:PBC_appendix_path_hist}
\end{figure}

\subsection{MSK-Chord}
\label{apd:MSK_data}
The MSK-CHORD dataset~\citep{jee2024automated} combines structured clinical data and natural language processing–derived annotations from 24,950 patients treated at Memorial Sloan Kettering Cancer Center. In this work, we focus on a breast cancer cohort.  After excluding individuals with substantial missing data, a total of 3,253 patients are retained for model training and evaluation. 

\paragraph{Static covariates.} We consider five static clinical covariates -- age, stage, hormone receptor (HR) status, HER2 status, and smoking history. The smoking history is included as a non-informative covariate to serve as a sanity check for the proposed explainability framework.

\paragraph{Longitudinal data.}  Two longitudinal tumor biomarkers are included:
\begin{itemize}
    \item CEA: carcinoembryonic antigen test in ng/ml,
    \item CA 15-3: cancer antigen 15-3 test in Units/ml.
\end{itemize}
These markers are routinely monitored in clinical practice and provide informative time-varying signals of disease progression. The average number of longitudinal measurements per patient is approximately 20.13.

\paragraph{Event definition.}
The event time is defined as the number of months from study registration to death.

\paragraph{Censoring.}
Patients who are alive at the end of follow-up are treated as right-censored. The dataset exhibits a high censoring rate of approximately $51\%$, which is typical for long-term observational oncology studies.

\paragraph{Supplementary figures.}
Additional illustrations are provided in Figure~\ref{fig:MSK_appendix_path_hist}, including example longitudinal trajectories up to the observed time for two selected patients and the empirical distribution of event times across the cohort.

\subsection{Summarize of datasets}
\label{appendix:data-summary}
\begin{table}[h]
\caption{Dataset characteristics: number of samples $(N)$, number of static features $(h)$, number of longitudinal features $(d)$, the average number of observations times over individuals $(\overline{m})$, the number of samples to be explained $(N_{\mathrm{exp}})$ and the censoring level $(\rho_c)$.}
\label{tab:summary_dataset}
\begin{center}
\begin{tabular}{lcccccc} 
 \toprule
 \textbf{Dataset} & $N$ & $N_{\mathrm{exp}}$ & $h$ & $d$ & $\overline{m}$ & $\rho_c$\\ 
 \midrule
 Simulation & 1000 & 200 & 3 & 5 & 56.86 & 2.6\%\\
 PBCseq & 312 & 150 & 2 & 5 & 6.52 & 57\%\\
 MSK-Chord & 3253 & 200 & 5 & 2 & 20.13 & 51\%\\
 \bottomrule
\end{tabular}
\end{center}
\label{tab:data_stat}
\end{table}

\begin{figure}[!h]
\centering
\includegraphics[width=1.\linewidth]{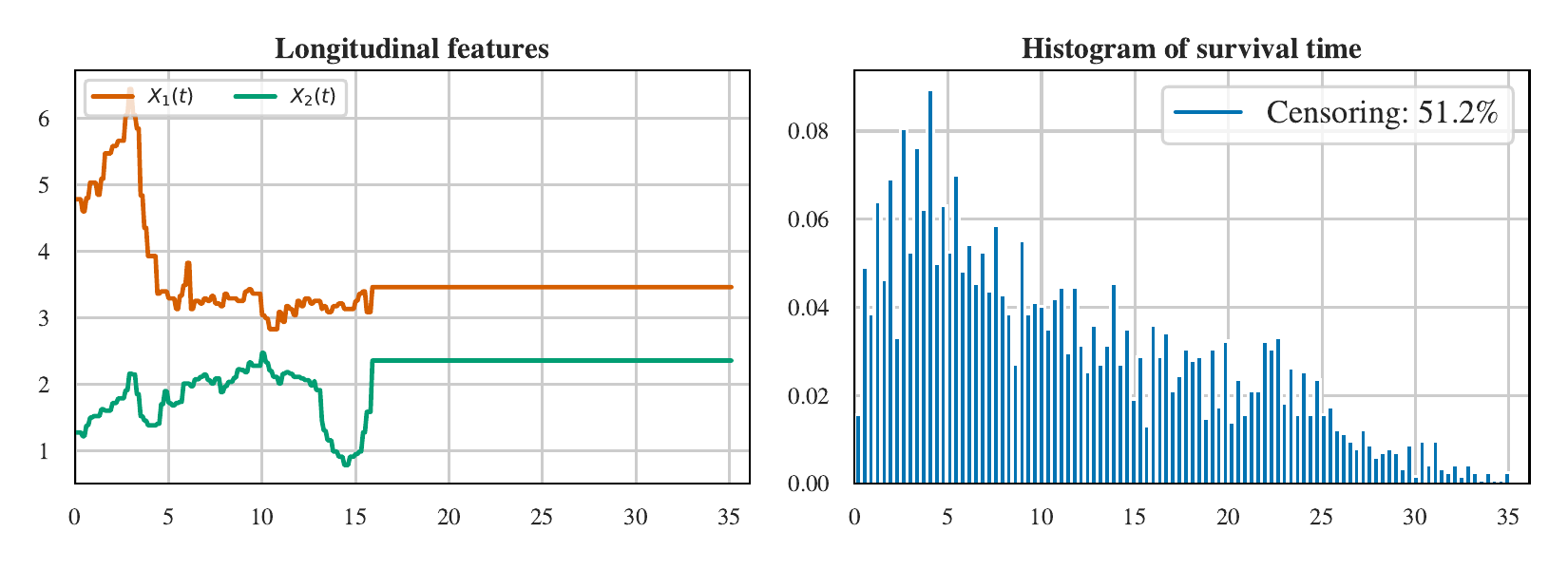}
\caption{MSK-Chord dataset -- Example longitudinal trajectories for two individuals, recorded up to their observed times (\textbf{left}, \textbf{middle}) and the Kaplan–Meier estimate of the event-time distribution (\textbf{right}).}
\label{fig:MSK_appendix_path_hist}
\end{figure}

\section{Algorithmic and Implementation Details}
\subsection{Description of predictive models}
\label{subappendix:pred_models}
\subsubsection{Data}
 \paragraph{Survival setting}
We consider a dataset of $N$ individuals,
\[
\mathcal{D}_n = \{ (\tilde{T}^i, \Delta^i, W^i, X^i) \}_{i=1}^N,
\]
where $\tilde{T}^i = \min(T^i, C^i)$ denotes the observed time, with $T^i$ the true event time and $C^i$ the censoring time. The indicator $\Delta^i = \mathds{1}\{T^i \leq C^i\}$ specifies whether the event of interest is observed. Static covariates are denoted by $W^i = \big(W^i_1, \ldots, W^i_h\big)\in \mathbb{R}^h$, while
\[
X^i = \big( X^i_1(t^i_1), \ldots, X^i_d(t^i_1), \ldots, X^i_1(t^i_{n_i}), \ldots, X^i_d(t^i_{n_i}) \big)
\]
represents $d$ longitudinal features measured at $n_i$ time points
$t^i_1 \leq \cdots \leq t^i_{n_i} \leq T^i$.

\subsubsection{CoxSig~\citep{bleistein2024dynamic}}
\paragraph{Modeling}
CoxSig proposes to parameterize the hazard function of the survival time $T^i$ via a latent state driven by a controlled differential equation,
\[
\lambda^i(t) = \exp\!\big( z^i(t) + \beta^\top W^i \big),
\qquad
\text{with } dz^i(t) = \mathbf{G}\!\big(z^i(t)\big)^\top dX^i(t)
\ \text{and } z^i(0) = 0,
\]
where the vector field $\mathbf{G} : \mathbb{R} \to \mathbb{R}^d$ and the parameter vector
$\beta \in \mathbb{R}^s$ are unknown.
Bleistein et al.~\citep{bleistein2024dynamic} show that this estimation problem can be linearized, yielding the following hazard representation:
\[
\lambda^i(t \mid X^i_{\le t}, W^i)
=
\exp\!\big(
\alpha^\top \mathbf{S}_N(X^i_{\le t}) + \beta^\top W^i
\big),
\]
where $\alpha$ and $\beta$ are learnable parameter vectors associated with longitudinal and static covariates, respectively. The term $\mathbf{S}_N(X^i_{\le t})$ denotes the truncated path signature of order $N \geq 1$ computed from the longitudinal trajectory observed up to time $t$~\citep{lyons2007differential, fermanian2021embedding}, providing a compact representation of temporal dynamics and feature interactions. The conditional survival probability is then given by
\[
\pi^i(t \mid t_{\mathrm{P}}; M)
=
\exp\!\left(
-\int_{t_{\mathrm{P}}}^{t} \lambda^i(u)\,du
\right).
\]

\paragraph{Loss function} Model parameters are estimated by minimizing a penalized negative log-likelihood with elastic-net regularization, 
\[\ell_{\textnormal{CoxSig}} = -\ell_{\textnormal{log-likelihood}} + \eta_1\textnormal{pen}_{\textnormal{EN}}(\alpha) + \eta_2\textnormal{pen}_{\textnormal{EN}}(\beta)\]
using a proximal gradient algorithm with backtracking line search~\citep{boyd2004convex}.

\paragraph{Hyperparameters} The regularization parameters $(\eta_1,\eta_2)$ and signature truncation order $N$ are selected by cross-validation using a combined criterion based on the C-index and Brier score. The search grid includes $\eta_1,\eta_2 \in \{1,e^{-1},\ldots,e^{-5}\}$ and $N \in \{2,3\}$; larger values of $N$ were excluded due to instability and degraded performance.

\subsubsection{Dynamic DeepHit~\citep{lee2019dynamic}}

DDH is a dynamical survival analysis algorithm that frames dynamical survival analysis as a classification problem. The time horizon $[0,\tau^{\text{max}}]$ is partitioned into $J$ contiguous intervals with cut-points $\{\tau_1,\ldots,\tau_J\}$. The network is then trained to predict a time interval of event for every individual, which is a multiclass classification task.  

\paragraph{Network Architecture.} Being adapted to competing events, Dynamic Deep-Hit combines a shared network with a cause-specific network. The \textit{shared network} is a combination of a RNN-like network that processes the longitudinal data and an attention mechanism, which helps the network decide which part of the history of the measurements is important. The \textit{cause-specific network} is a feed-forward network taking as an input the history of embedded measurements and learning a cause-specific representation. See Figure \ref{fig:ddh} for a graphical representation of the network's structure. 
\begin{figure}[h!]
    \centering
    \includegraphics[width=0.5\textwidth]{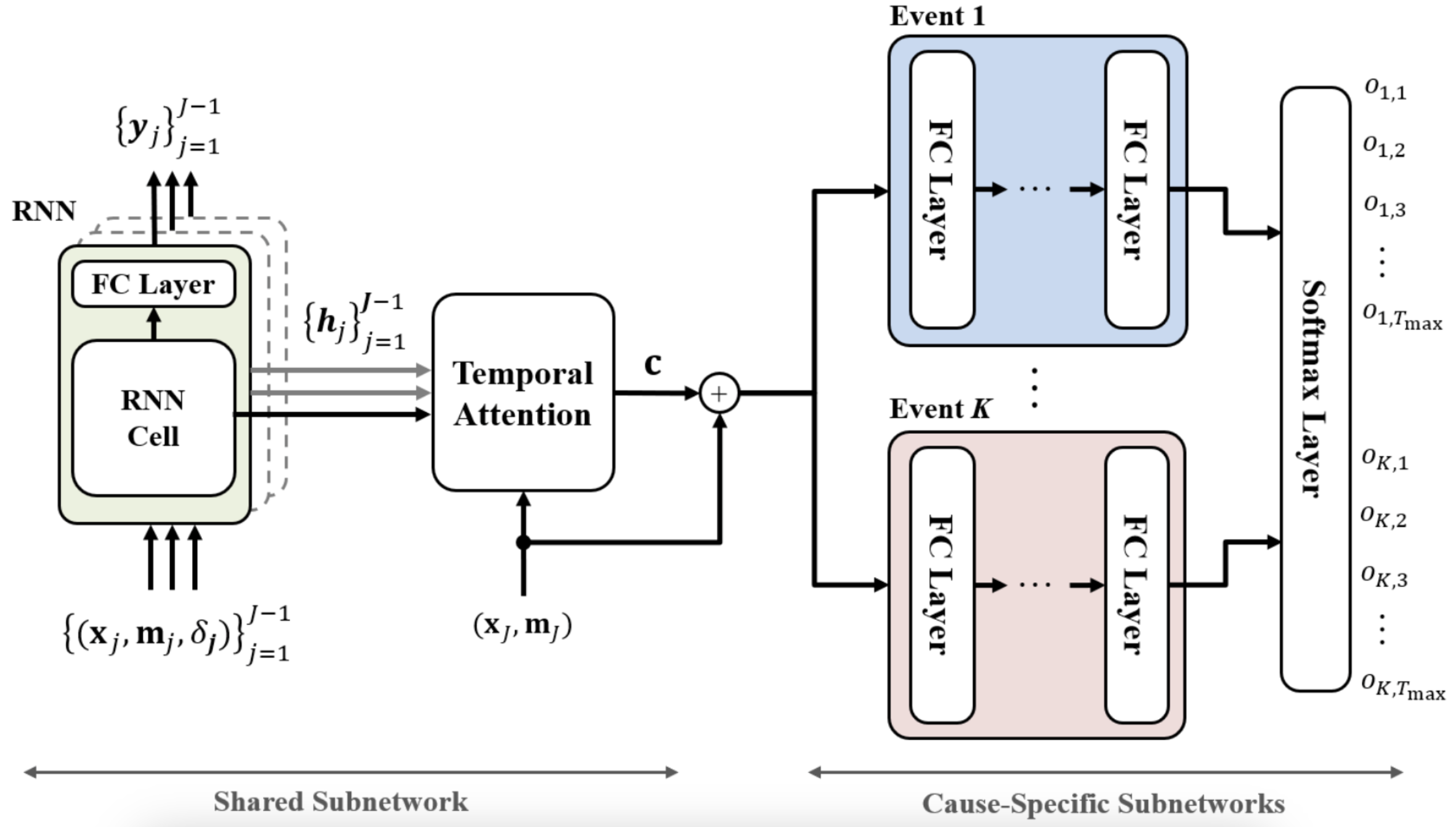}
    \caption{Network structure of DDH. Figure is taken from \citet{lee2019dynamic}.}
    \label{fig:ddh}
\end{figure}

For simplicity, we present the single-event setting. Let $0=\tau_0 < \tau_1 < \cdots < \tau_J$ denote a discretization of time into $J$ intervals $(\tau_{j-1}, \tau_j]$. 
For an individual $i$ with static covariates $W^i$ and longitudinal observations available up to a prediction time $t_{\mathrm{P}}$, the DDH model outputs a discrete-time event distribution over the intervals:
\[
o^i_j = \mathds{P}\!\left( T^i \in (\tau_{j-1}, \tau_j] \,\middle|\, X^i_{\le t_{\mathrm{P}}}, W^i \right),
\qquad j=1,\ldots,J.
\]
The corresponding (unconditional) discrete survival function is
\[
S^i(\tau_j \mid t_{\mathrm{P}}; M) = \mathds{P}(T^i > \tau_j \mid X^i_{\le t_{\mathrm{P}}}, W^i; M)
= 1 - \sum_{k=1}^{j} o^i_k.
\]
Finally, the conditional survival probability given survival up to $t_{\mathrm{P}}$ is obtained by normalization:
\[
\pi^i(t \mid t_{\mathrm{P}}; M)
=
\mathds{P}(T^i \ge t \mid T^i \ge t_{\mathrm{P}}, X^i_{\le t_{\mathrm{P}}}, W^i; M)
=
\frac{S^i(t \mid t_{\mathrm{P}}; M)}{S^i(t_{\mathrm{P}} \mid t_{\mathrm{P}}; M)}.
\]

\paragraph{Loss Function.} The loss function of DDH is a sum of three loss functions 
\[
\ell_{\textnormal{DDH}} = \ell_{\textnormal{log-likelihood}} + \ell_{\textnormal{ranking}} + \ell_{\textnormal{prediction}}. 
\]

The first loss maximizes the conditional likelihood of dying in the interval $[t_{k},t_{k+1}[$ given that the individual has survived up to time $t_k$. The second loss favors correct rankings among at risk individuals: an individual experiencing an event at time $T^i$ should have a higher risk score at time $t < T^i$ than an individual $j$ for which $T^j > T^i$. The third loss is a prediction loss, which measures the difference between the value of the time-dependent features and a prediction of this value made by the shared network. The loss is minimized using Adam \citep{kingma2014adam}.  

\paragraph{Hyperparameters.} In our setting, we use the network in its original structure. The learning rate is set to $e^{-4}$ and the number of epochs to 300. 

\subsection{Time-dependent C-index and Brier score}\label{subappendix:metrics}
We evaluate predictive performance using time-dependent metrics based on the individual-specific conditional survival probability
$\pi^i(t_{\text{P}} + \delta t \mid t_{\text{P}}; M)$,
which is the probability that individual $i$ survives beyond $t_{\text{P}} + \delta t$, conditional on having survived up to the prediction time $t_{\text{P}}$, as estimated by model $M$.

\paragraph{Time-dependent Concordance Index.} Following \citet{lee2019dynamic}, we measure the discriminative power of our models by using a time-dependent concordance index $C(t_{\text{P}},\delta t)$ that captures our models ability to correctly rank individuals on their predicted probability of survival. The concordance index $C(t_{\text{P}},\delta t)$ is then finally computed as  

\begin{align*}
\frac{\sum\limits_{j=1}^n \sum\limits_{i=1}^n\mathds{1}_{\pi^i(t_{\text{P}} + \delta t \mid t_{\text{P}}; M) > \pi^j(t_{\text{P}} + \delta t \mid t_{\text{P}}; M)} \mathds{1}_{\tilde{T}^i > \tilde{T}^j,\, \tilde{T}^j \in [t_{\text{P}},t_{\text{P}} + \delta t], \Delta^j = 1}}{\sum\limits_{j=1}^n \sum\limits_{i=1}^n \mathds{1}_{\tilde{T}^i > \tilde{T}^j,\, \tilde{T}^j \in [t_{\text{P}},t_{\text{P}} + \delta t], \Delta^j = 1}}.
\end{align*}
This metric captures the capacity of our model to discriminate between $j$ and another individual $i$ through the conditional probability of survival. 

\paragraph{Brier Score.} While the concordance index is a ranking-based measure, the Brier Score measures the accuracy in predictions by comparing the estimated survival function and the survival indicator function \citep{lee2019dynamic,kvamme2019time,kvamme2023brier}. Formally, we define the Brier score $\textnormal{BS}(t_{\text{P}},\delta t)$ as
\begin{align*}
    \frac{1}{n}\sum\limits_{i=1}^n \mathds{1}_{\tilde{T}^i \leq t_{\text{P}} + \delta t, \Delta^i=1} [\pi^i(t_{\text{P}} + \delta t \mid t_{\text{P}}; M)]^2
    +  \frac{1}{n}\sum\limits_{i=1}^n \mathds{1}_{\tilde{T}^i > t_{\text{P}} + \delta t} [1-\pi^i(t_{\text{P}} + \delta t \mid t_{\text{P}}; M)]^2.
\end{align*}
Contrarily to the C-index, the Brier score is a measure of calibration of the predictions: it measures the distance between the estimated survival function and the indicator function of survival on the interval $[t_{\text{P}},t_{\text{P}}+\delta t]$.


\section{Supplementary materials for Section~\ref{sec:results}}
\subsection{Optimal number of evaluated coalitions with Dynamic WindowSHAP}
\label{apd:optimalcoaltions}
Due to the large number of coalitions generated by Dynamic WindowSHAP and the resulting memory constraints, we restrict the analysis to a subset of coalitions. To balance computational cost and explanation quality, we conduct an experiment on the simulated dataset with varying numbers of evaluated coalitions.

As shown in Figure~\ref{fig:optimalcoaltions}, using 50 coalitions achieves explanation performance comparable to 100, while 25 leads to a noticeable degradation. We therefore select 50 as an efficient trade-off between accuracy and runtime.

\begin{figure}[!h]
    \centering
    \includegraphics[width=\linewidth]{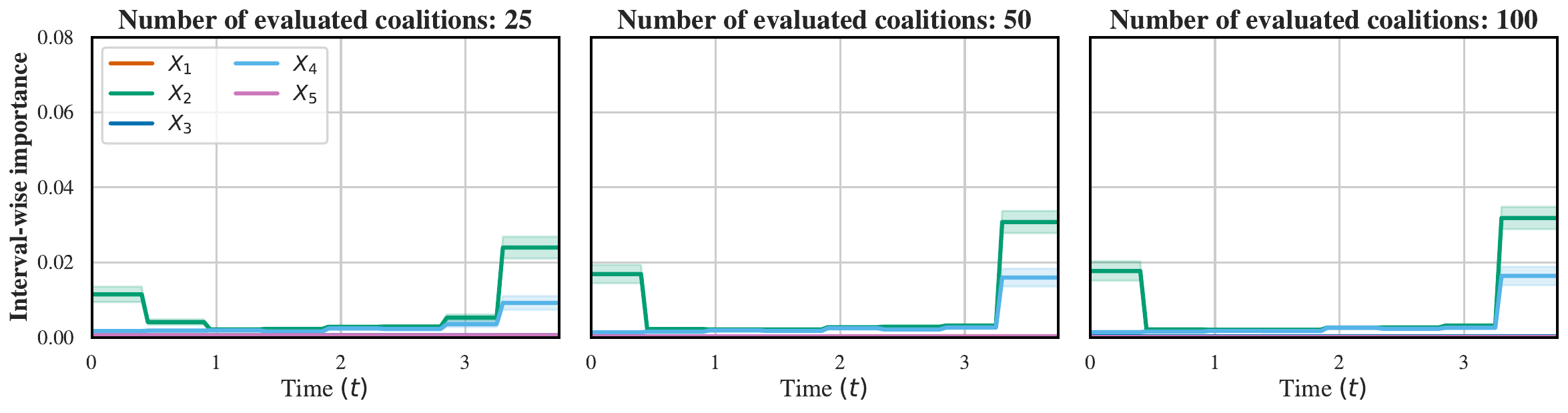}
    \caption{Different number of evaluated coalitions on simulated dataset}
    \label{fig:optimalcoaltions}
\end{figure}

\subsection{Predictive performance on the real datasets}
\label{apd:additional_pred_perf}
We report here the predictive performance of CoxSig, DDH and DySurv on the simulated dataset (Figure~\ref{fig:simu_perf}), the PBCseq dataset (Figure~\ref{fig:PBC_perf}) and the MSK-Chord dataset (Figure~\ref{fig:MSK_perf}).

\begin{figure}[!h]
    \centering
    \includegraphics[width=\linewidth]{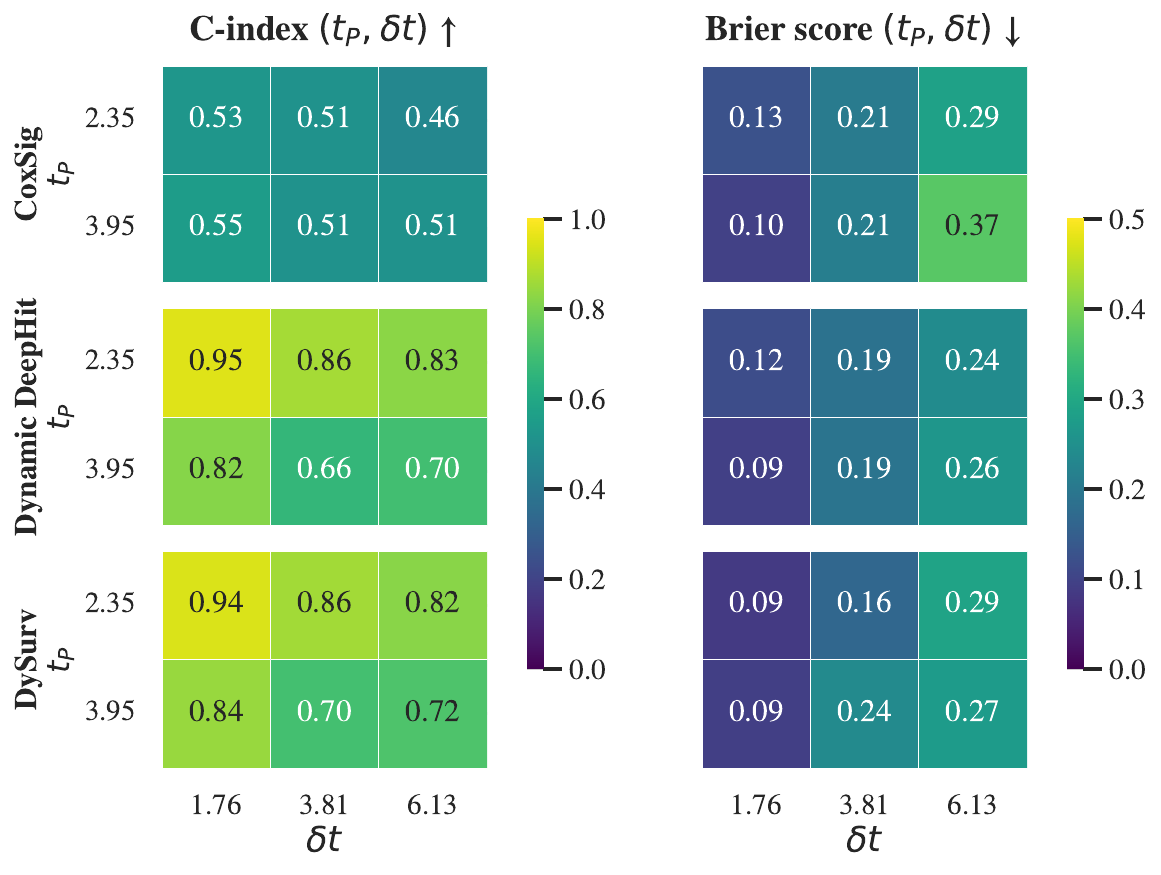}
    \caption{PBCseq dataset -- Predictive performance. C-Index (\textbf{left}) and Brier score (\textbf{right}) of the three models: CoxSig (\textbf{top}), DDH (\textbf{middle}) and DySurv (\textbf{bottom}) evaluated at chosen points $(t_{\text{P}},\delta t)$.}
    \label{fig:PBC_perf}
\end{figure}

\begin{figure}[!h]
    \centering
    \includegraphics[width=\linewidth]{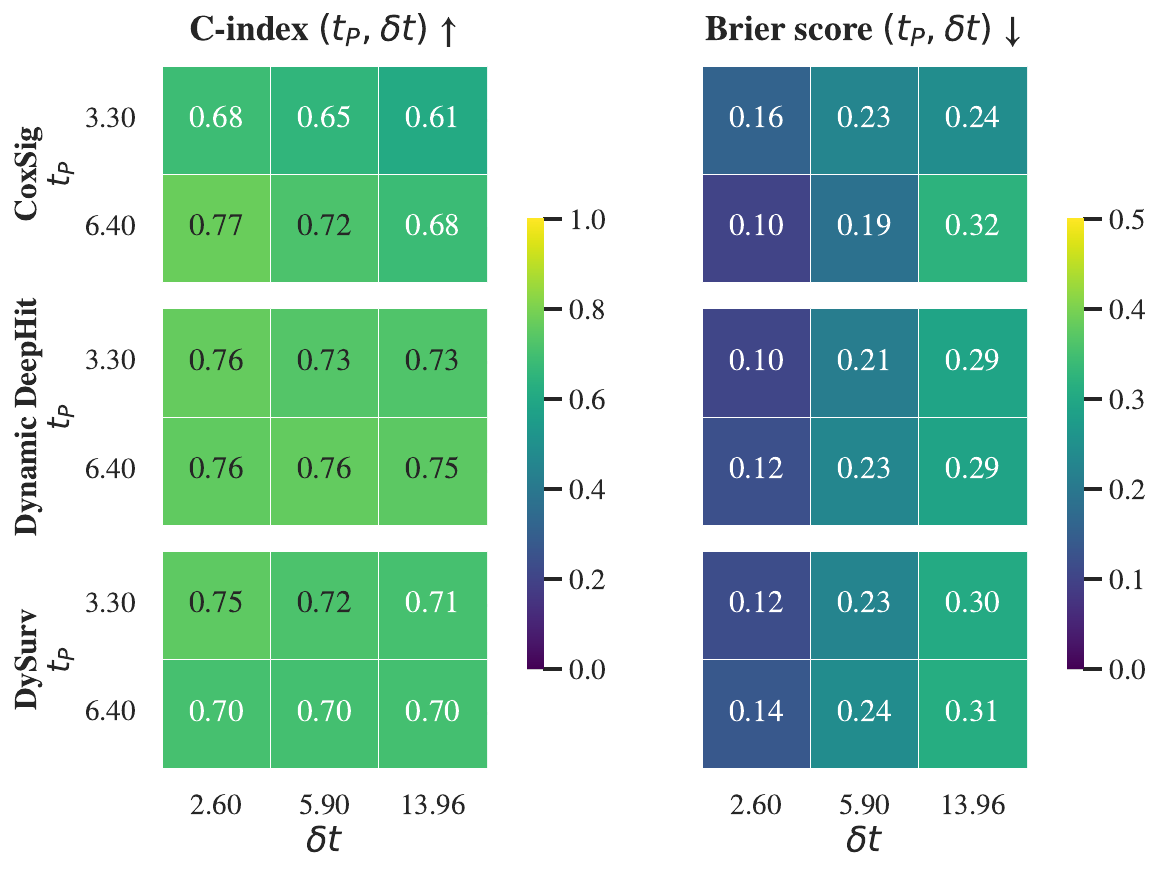}
    \caption{MSK-Chord dataset -- Predictive performance. C-Index (\textbf{left}) and Brier score (\textbf{right}) of the three models: CoxSig (\textbf{top}), DDH (\textbf{middle}) and DySurv (\textbf{bottom}) evaluated at chosen points $(t_{\text{P}},\delta t)$.}
    \label{fig:MSK_perf}
\end{figure}

\newpage
\subsection{Explanation performance in dynamic prediction setting (Expected remaining restricted survival time)}

We provide additional explanation results for PBCseq and MSK-Chord using the expected remaining restricted survival time as the prediction target.
\label{apd:additional_expl_perf}
\begin{figure}[h]
    \centering
    \includegraphics[width=\linewidth]{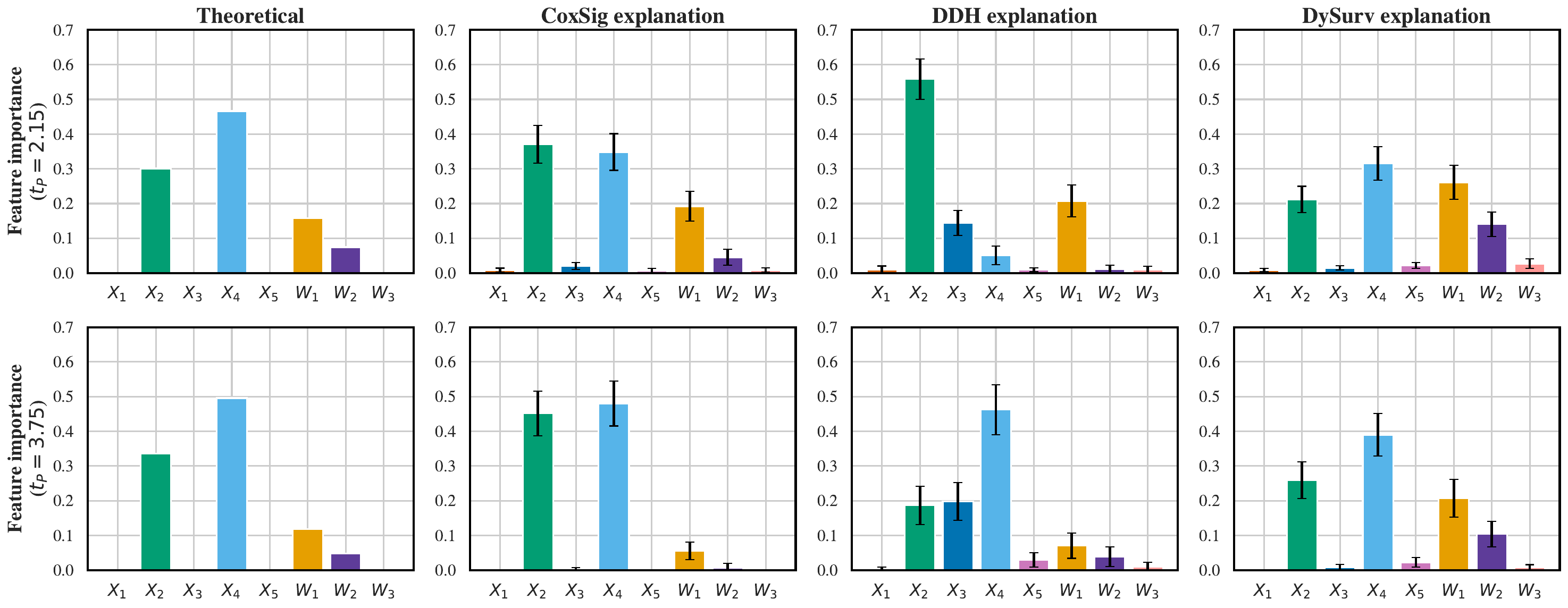}
     \caption{Simulated dataset -- \textbf{Feature importance}. Theoretical values (\textbf{left}) and estimates for \texttt{DynSurvX} for CoxSig, DDH, and DySurv (\textbf{three right panels}) at two prediction times $t_{\text{P}} = 2.15$ (\textbf{top}) and $t_{\text{P}} = 3.75$ (\textbf{bottom}), with the red vertical line separates longitudinal features from static features.} 
    \label{fig:simu_shap_feat}
\end{figure}

\begin{figure}[h]
    \centering
    \includegraphics[width=\linewidth]{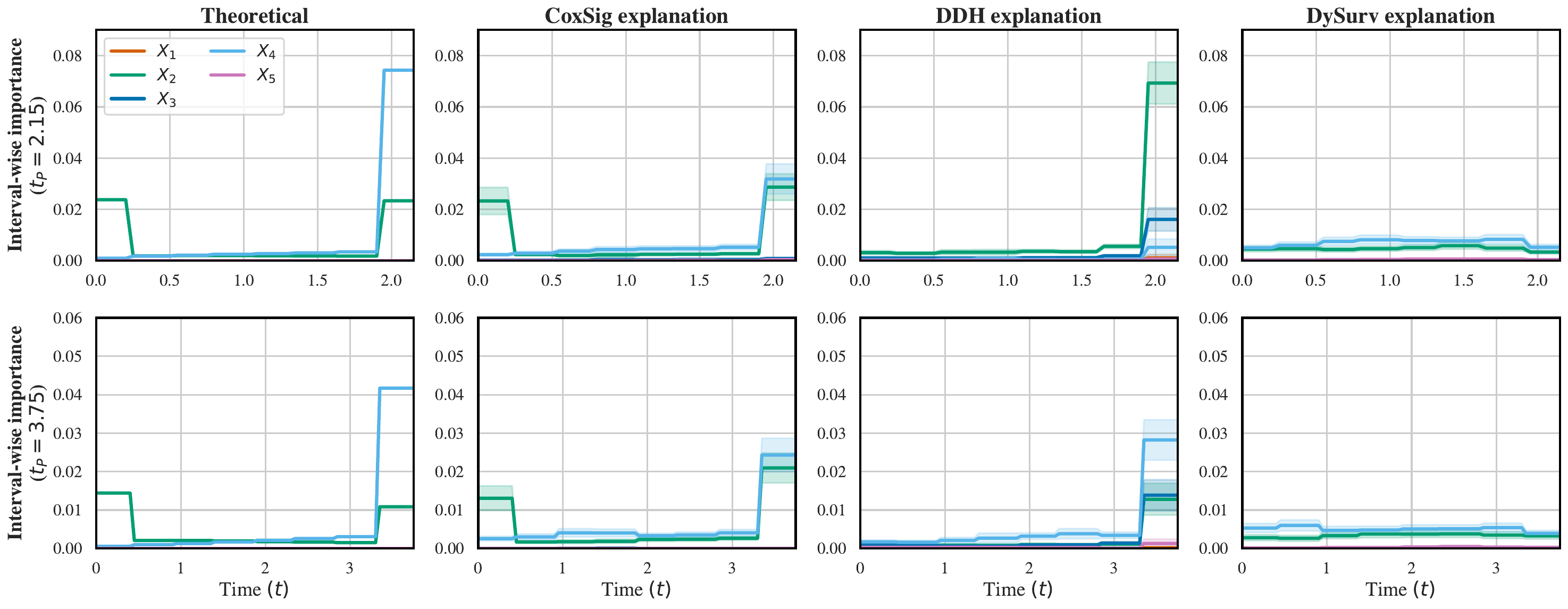}
    \caption{Simulated dataset -- \textbf{Interval-wise feature importance}. Theoretical values (\textbf{left}) and estimates for \texttt{DynSurvX} for CoxSig, DDH, and DySurv (\textbf{three right panels}) at two prediction times $t_{\text{P}} = 2.15$ (\textbf{top}) and $t_{\text{P}} = 3.75$ (\textbf{bottom}).}
    \label{fig:simu_shap_longi}
\end{figure}

\label{apd:additional_expl_perf}
\begin{figure}[h]
    \centering
    \includegraphics[width=\linewidth]{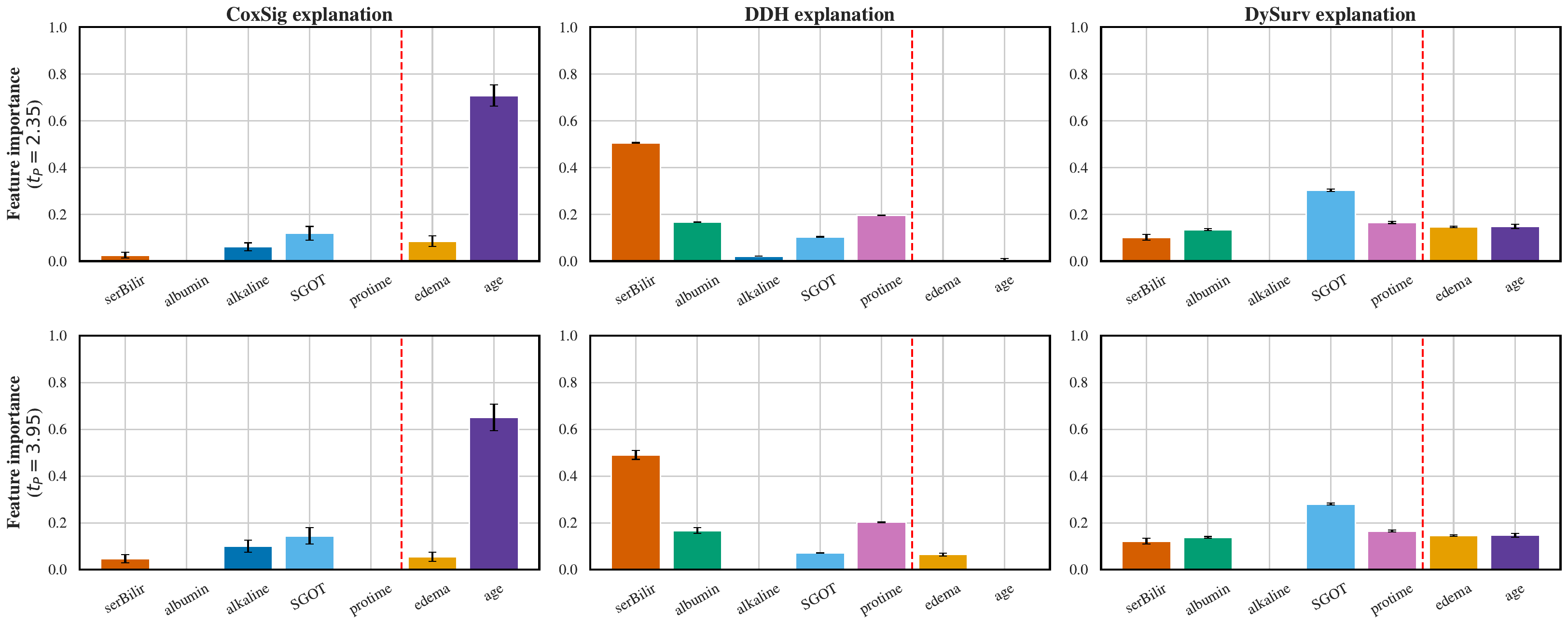}
     \caption{PBCseq dataset -- \textbf{Feature importance} comparison between estimates obtained with \texttt{DynSurvX} for CoxSig (\textbf{left}), DDH (\textbf{middle}) and DySurv (\textbf{right}) at two prediction times $t_{\text{P}} = 2.35$ (\textbf{top}) and $t_{\text{P}} = 3.95$ (\textbf{bottom}), with the red vertical line separates longitudinal features from static features.} 
    \label{fig:PBC_shap_feat}
\end{figure}

\begin{figure}[h]
    \centering
    \includegraphics[width=\linewidth]{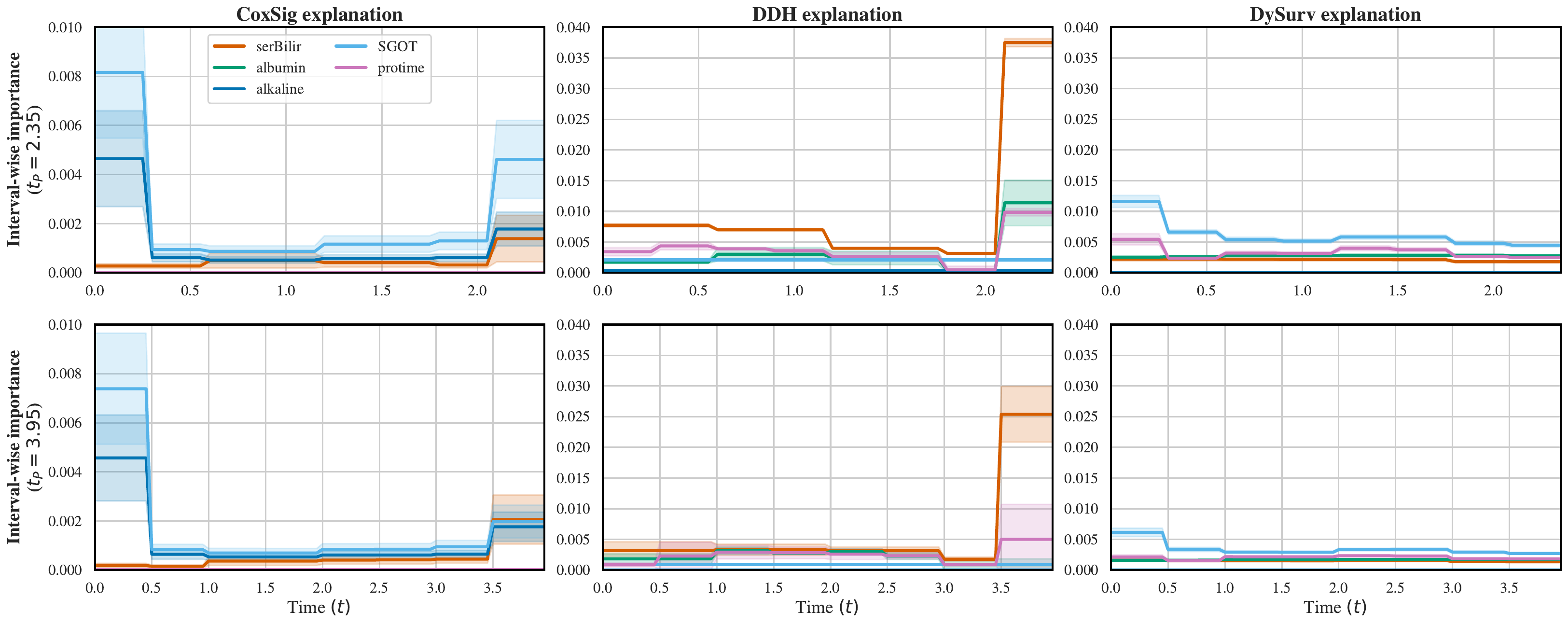}
    \caption{PBCseq dataset -- \textbf{Interval-wise feature importance} comparison between estimates obtained with \texttt{DynSurvX} for CoxSig (\textbf{left}), DDH (\textbf{middle}) and DySurv (\textbf{right}) at two prediction times $t_{\text{P}} = 2.35$ (\textbf{top}) and $t_{\text{P}} = 3.95$ (\textbf{bottom}).}
    \label{fig:PBC_shap_longi}
\end{figure}

\begin{figure}[!h]
    \centering
    \includegraphics[width=\linewidth]{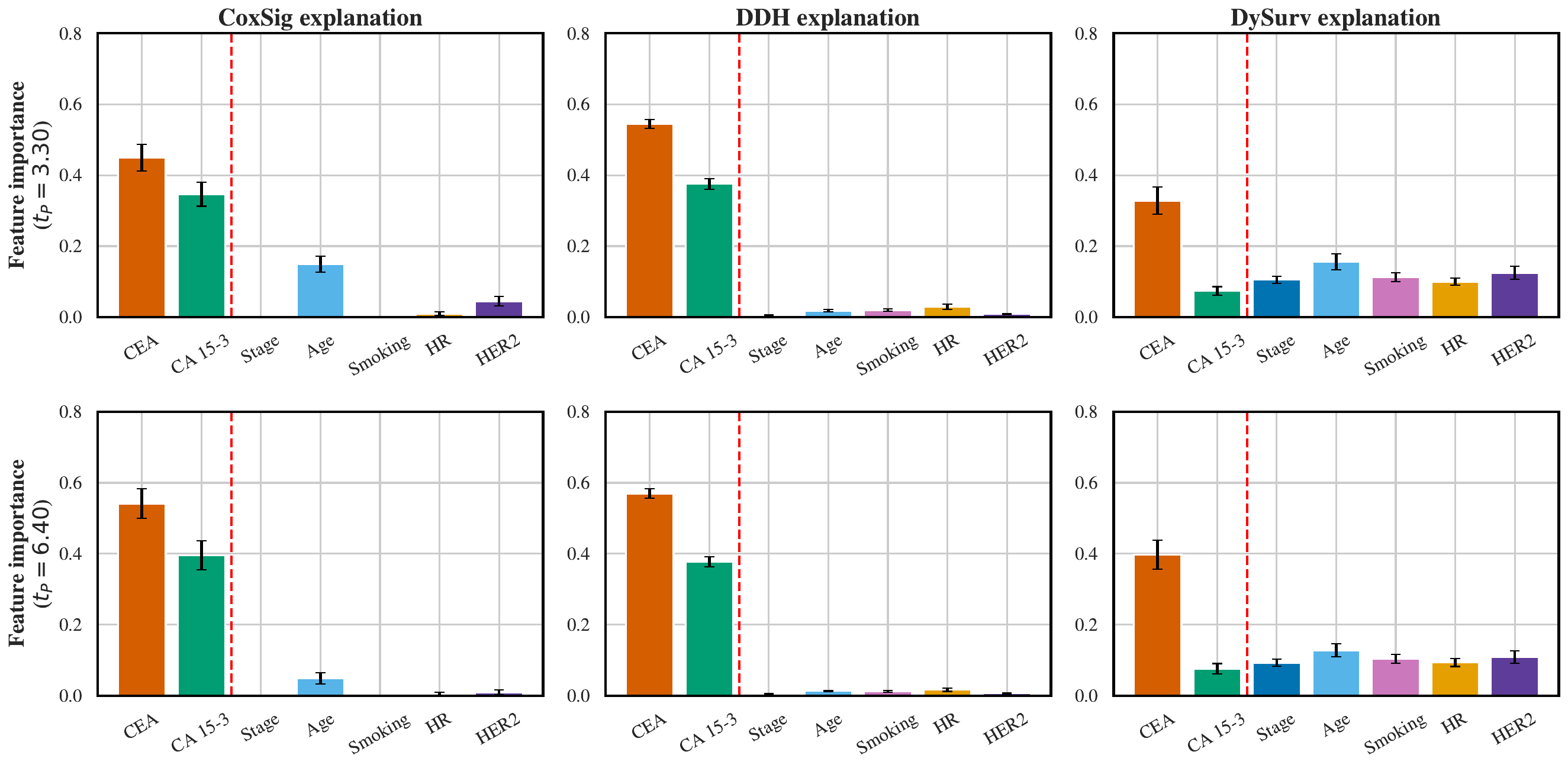}
    \caption{MSK-Chord dataset -- \textbf{Feature importance} comparison between estimates obtained with \texttt{DynSurvX} for CoxSig (\textbf{left}), DDH (\textbf{middle}) and DySurv (\textbf{right}) at two prediction times $t_{\text{P}} = 3.3$ (\textbf{top}) and $t_{\text{P}} = 6.4$ (\textbf{bottom}), with the red vertical line separates longitudinal features from static features.}
    \label{fig:MSK_shap_feat}
\end{figure}

\begin{figure}[!h]
    \centering  \includegraphics[width=\linewidth]{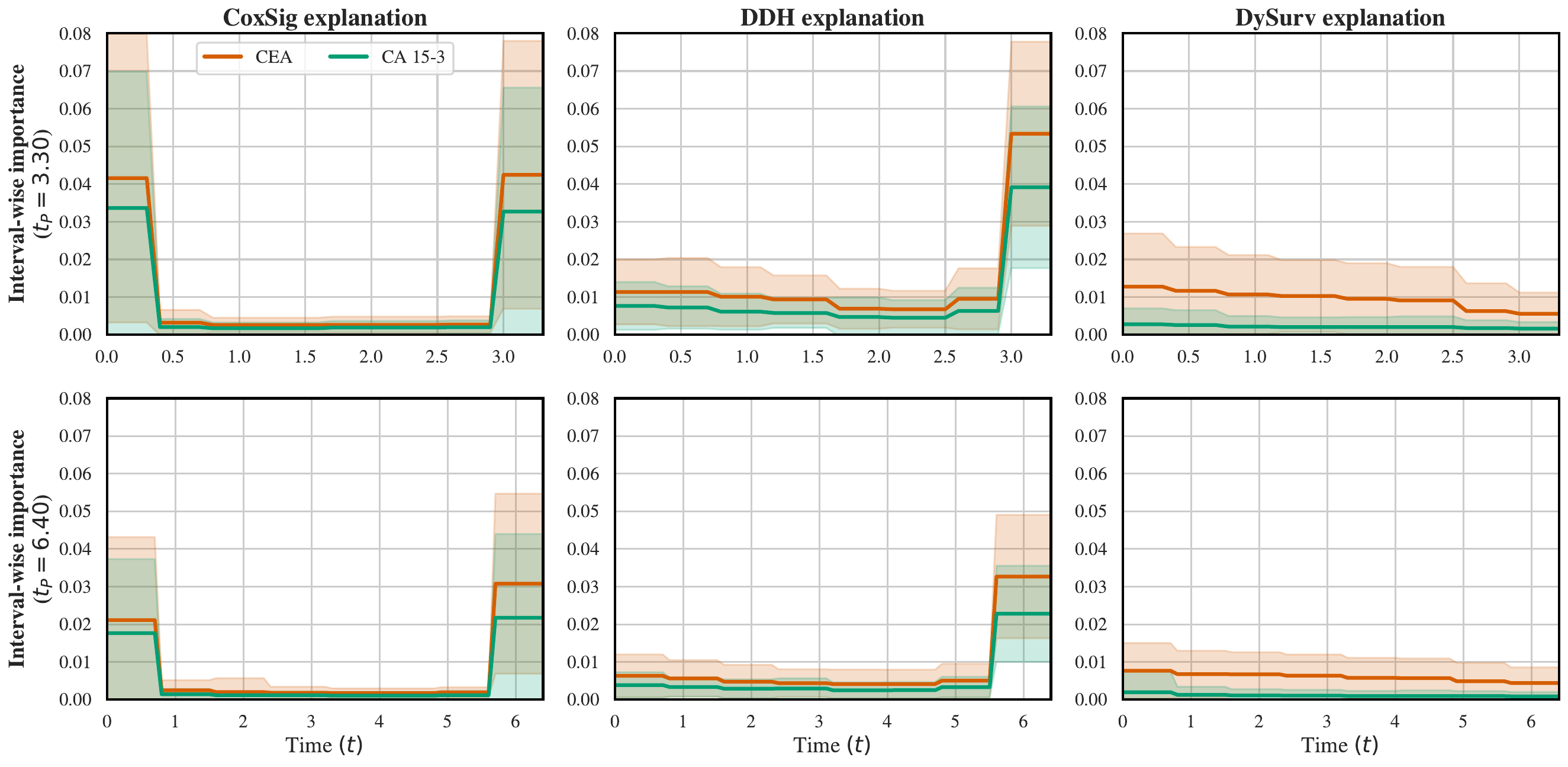}
    \caption{MSK-Chord dataset -- \textbf{Interval-wise feature importance} comparison between estimates obtained with \texttt{DynSurvX} for CoxSig (\textbf{left}), DDH (\textbf{middle}) and DySurv (\textbf{right}) at two prediction times $t_{\text{P}} = 3.3$ (\textbf{top}) and $t_{\text{P}} = 6.4$ (\textbf{bottom}).}
    \label{fig:MSK_shap_longi}
\end{figure}

\subsection{Optimal $\delta_1$} 
\label{apd:optimaldelta}

The threshold $\delta_1$ controls which longitudinal features are considered informative and therefore eligible for temporal splitting in Phase~2 of \texttt{DynSurvX}. Its choice is critical for balancing explanation fidelity and computational efficiency.

Importantly, in our framework the choice of $\delta_1$ relies entirely on the feature-level importance scores computed in Phase~1. As a result, Phase~1 must be executed prior to Phase~2 to obtain normalized importance scores for all longitudinal features at the individual level. These scores provide a coarse but essential screening step that identifies which features warrant further temporal refinement.

To select $\delta_1$, we adopt an \emph{elbow method} based on the distribution of feature-level importance scores obtained in Phase~1. For a fixed prediction time and model, we compute the normalized importance scores of all longitudinal features for each individual and sort them in descending order. This yields a decreasing curve that reflects how explanatory mass is distributed across features. We then average these sorted curves across individuals to obtain a smooth, population-level importance profile.

The optimal value of $\delta_1$ is chosen at the first large drop in this averaged curve, corresponding to the point beyond which additional features contribute only marginally to the prediction. This choice retains the most influential longitudinal features while filtering out weak or noisy contributors.

Figure~\ref{fig:opt_threshold} illustrates this procedure on the simulated dataset and PBCseq dataset. Orange curves correspond to individual-level sorted importance profiles, while the green curve shows their average. The elbow point of the mean curve defines the threshold $\delta_1$.

\begin{figure}[!h]
    \centering
    \begin{minipage}{0.48\linewidth}
    \includegraphics[width=\linewidth]{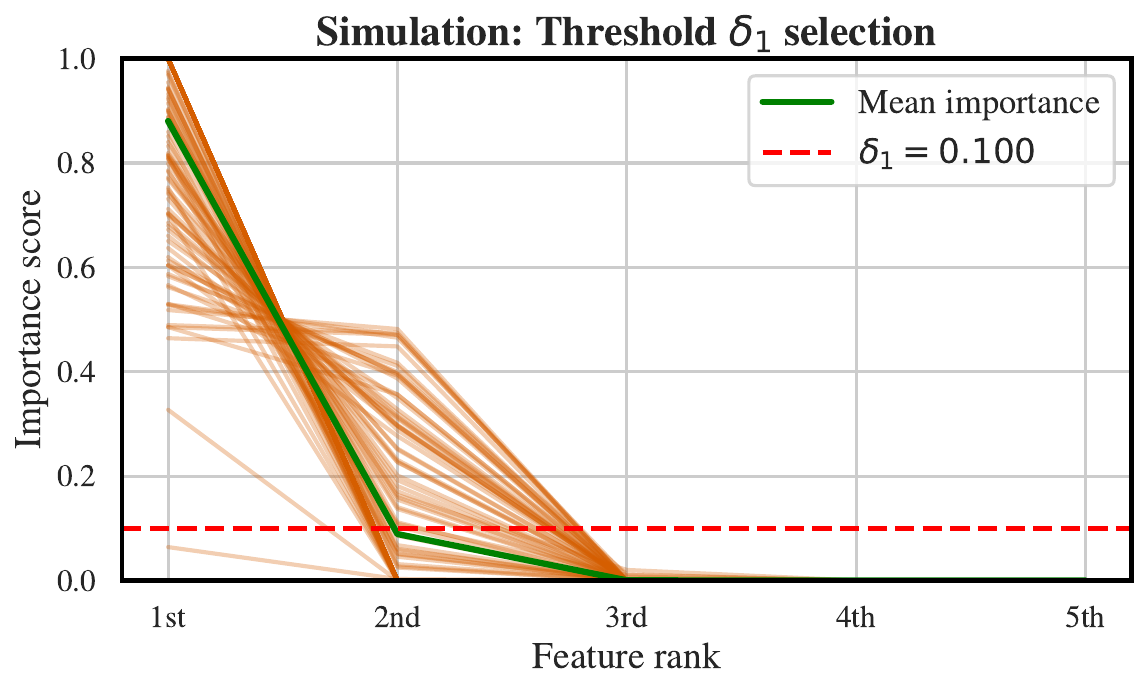}
    \end{minipage}
    \begin{minipage}{0.48\linewidth}
    \includegraphics[width=\linewidth]{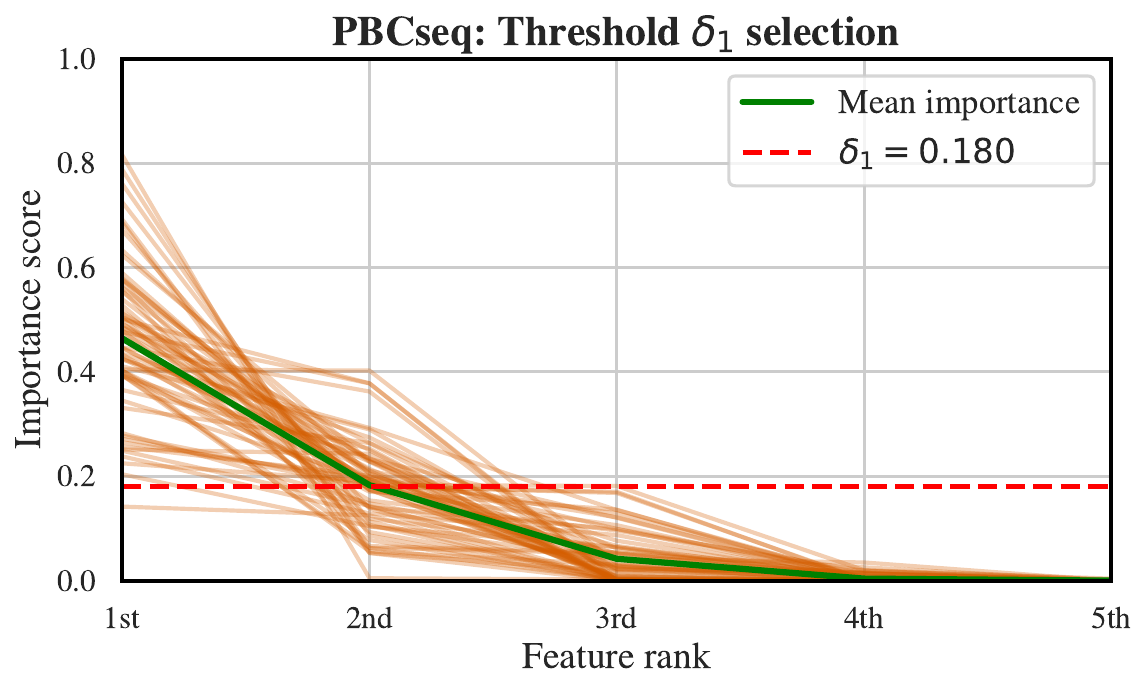}
    \end{minipage}
    \caption{Elbow-based selection of $\delta_1$ on simulated dataset (\textbf{left}) and PBC dataset (\textbf{right}).  Individual-level sorted longitudinal feature importance scores (\textbf{orange}) and their population average (\textbf{green}). The elbow point of the mean curve defines the threshold $\delta_1$.}
    \label{fig:opt_threshold}
\end{figure}

\subsection{Median survival time}
\label{apd:additional_exp}

Given the conditional survival probability
$\pi^i(t \mid t_{\mathrm{P}}; M)$ in Section~\ref{sec:background}, the median residual survival time is defined as
\[m\!\left[T^i - t_{\mathrm{P}} \,\middle|\, W^i, X^i_{\le t_{\mathrm{P}}}; M\right]= \inf\{ t - t_{\text{P}} : \pi^i(t \mid t_{\text{P}}; M) > 0.5\}.\]
which provides a robust summary of the conditional survival distribution and is less sensitive to outliers and skewness than the expected remaining restricted survival time.

Across all datasets, \texttt{DynSurvX} yields consistent global and temporal explanations when using the median residual survival time instead of the expected remaining restricted survival time as the prediction target. Both feature-value attribution and interval-wise importance remain stable across these two targets, indicating that the proposed method is robust to the choice of survival summary (see  Figures~\ref{fig:simu_shap_feat_med},~\ref{fig:simu_shap_longi_med},~\ref{fig:PBC_shap_feat_med},~\ref{fig:PBC_shap_longi_med},~\ref{fig:MSK_shap_feat_med} and~\ref{fig:MSK_shap_longi_med}).
 
\begin{figure}[!h]
\centering    
\includegraphics[width=\linewidth]{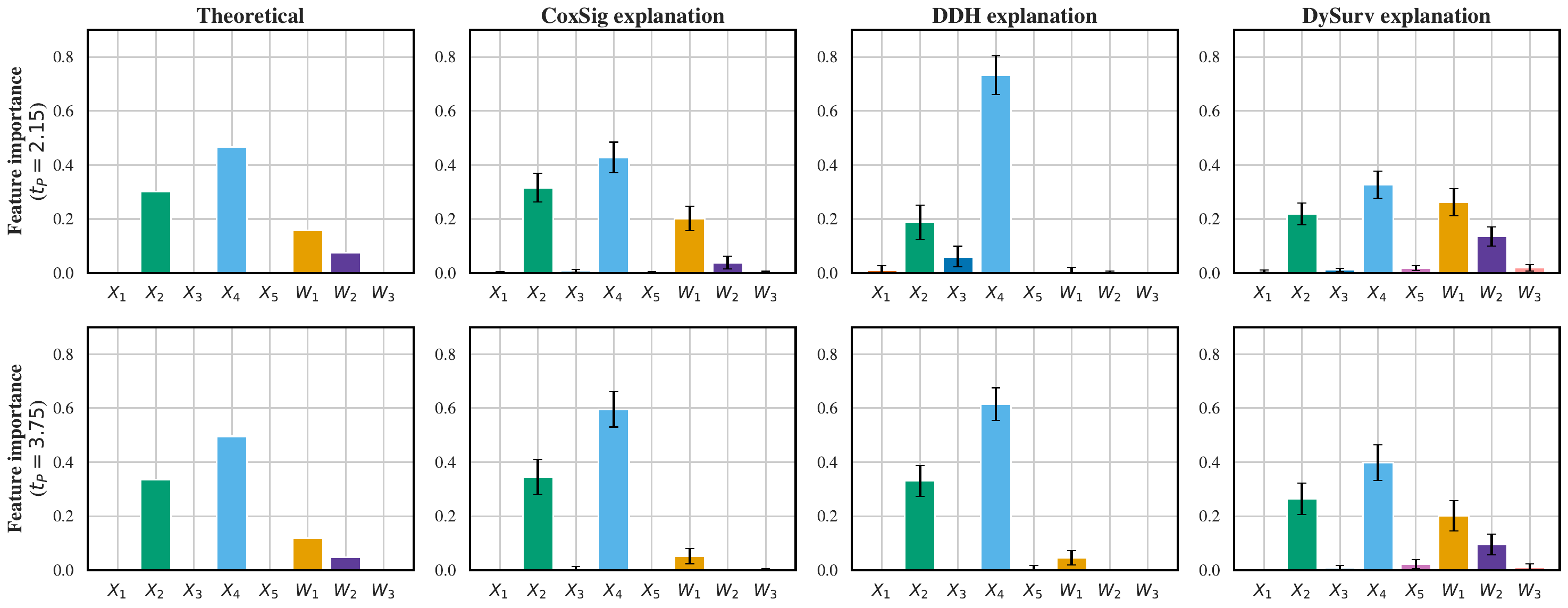}
\caption{Simulated dataset + median survival time as prediction function -- \textbf{Feature importance} comparison between the theoretical values (\textbf{left}) and estimates obtained with \texttt{DynSurvX}, together with 95\% variability intervals shown as error bars, for CoxSig (\textbf{middle}) and DDH (\textbf{right}) at two prediction times $t_{\text{P}} = 2.15$ (\textbf{top}) and $t_{\text{P}} = 3.75$ (\textbf{bottom}).}
\label{fig:simu_shap_feat_med}
\end{figure}

\begin{figure}[!h]
\centering    
\includegraphics[width=\linewidth]{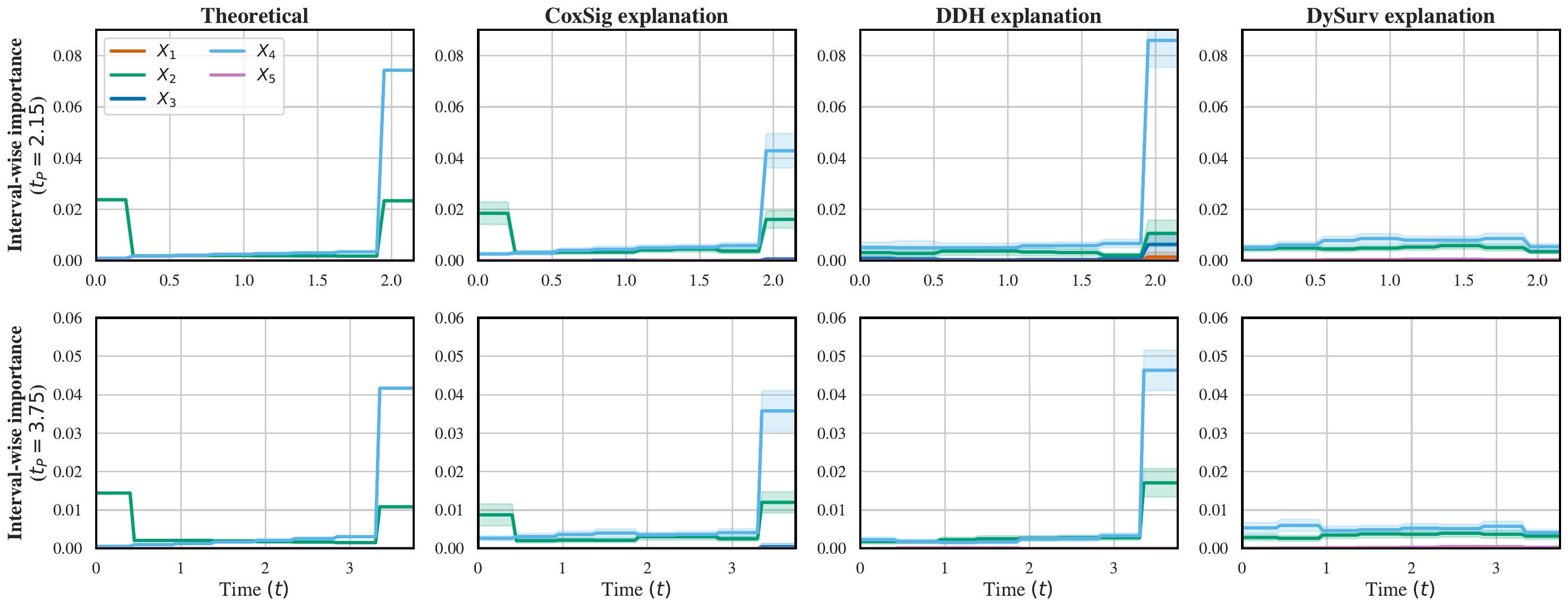}
    \caption{Simulated dataset + median survival time as prediction function -- \textbf{Interval-wise feature importance} comparison between the theoretical values (\textbf{left}) and estimates obtained with \texttt{DynSurvX}, together with 95\% variability intervals, for CoxSig, DDH, and DySurv (\textbf{three right panels}) at two prediction times $t_{\text{P}} = 2.15$ (\textbf{top}) and $t_{\text{P}} = 3.75$ (\textbf{bottom}).}
\label{fig:simu_shap_longi_med}
\end{figure}

\begin{figure}[!h]
    \centering
    \includegraphics[width=\linewidth]{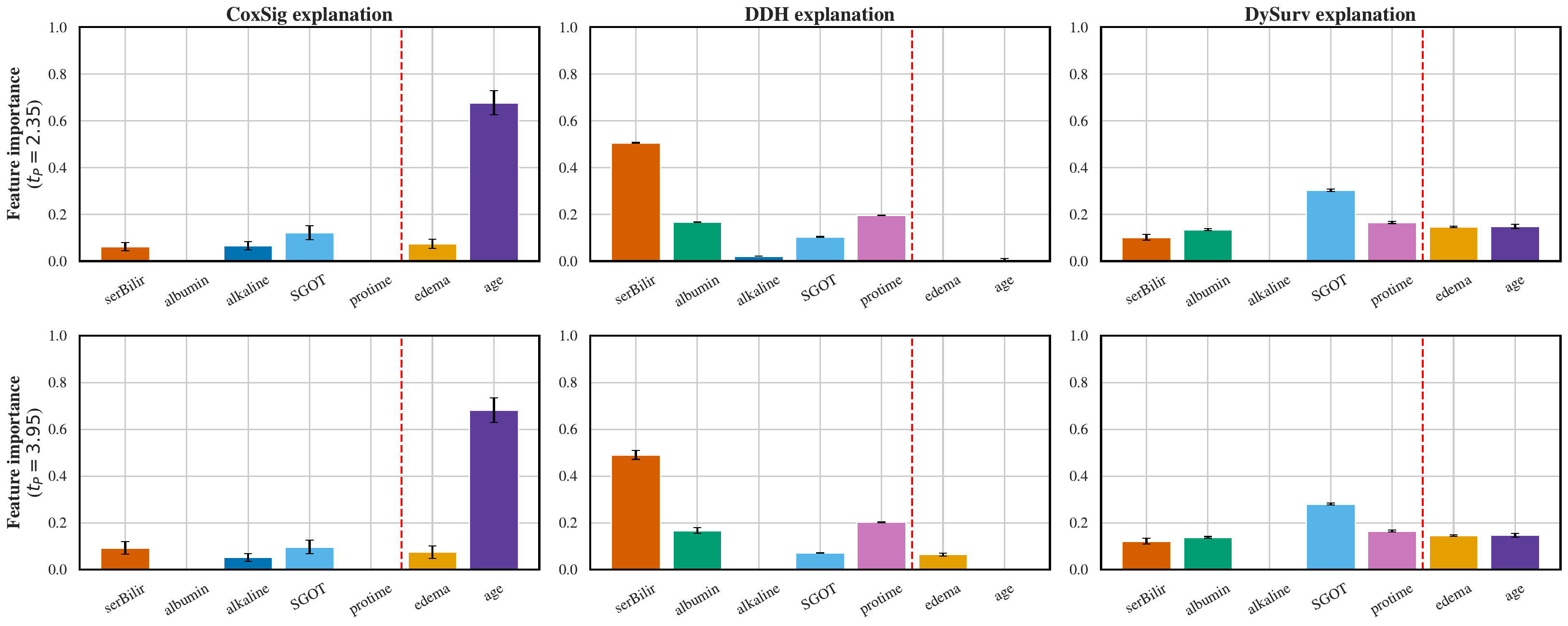}
    \caption{PBCseq dataset + median survival time as prediction function -- \textbf{Feature importance} comparison between estimates obtained with \texttt{DynSurvX}, together with 95\% variability intervals shown as error bars, for CoxSig, DDH, and DySurv (\textbf{three right panels}) at two prediction times $t_{\text{P}} = 2.35$ (\textbf{top}) and $t_{\text{P}} = 3.95$ (\textbf{bottom}).} 
    \label{fig:PBC_shap_feat_med}
\end{figure}

\begin{figure}[!h]
    \centering
    \includegraphics[width=\linewidth]{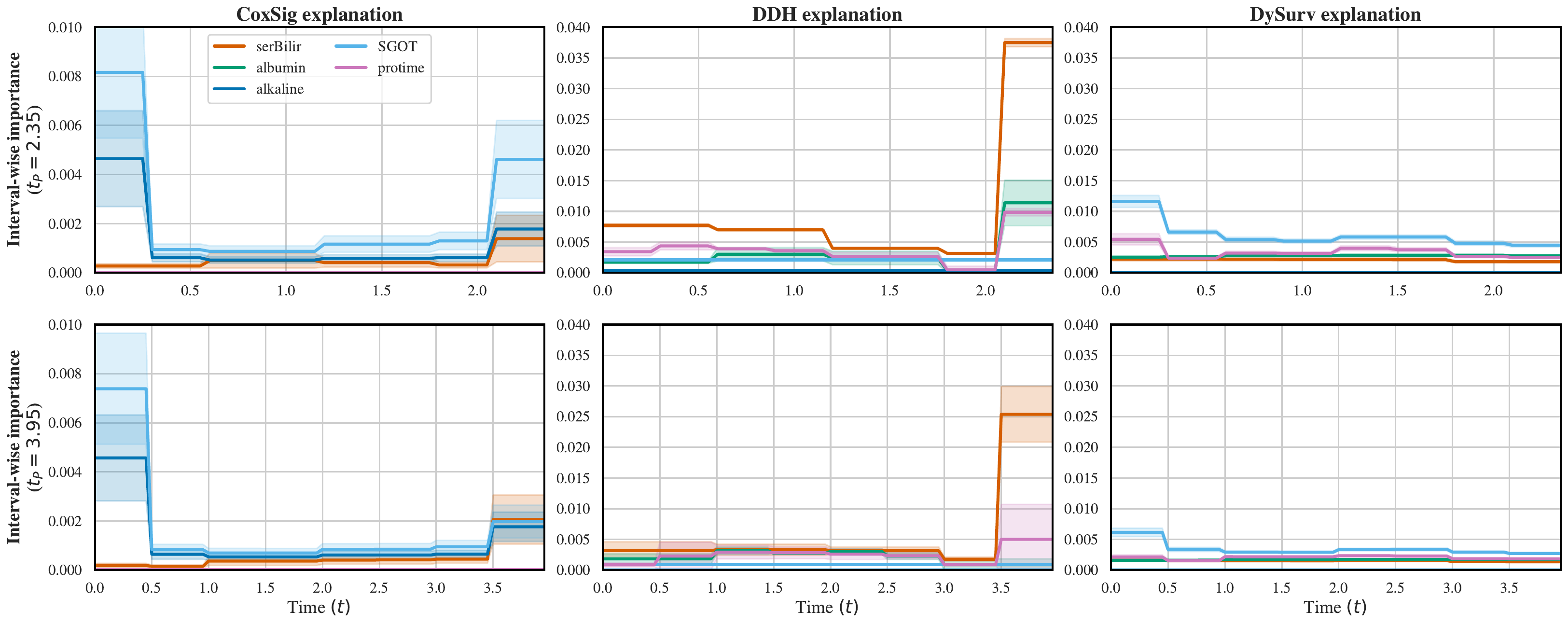}
    \caption{PBCseq dataset + median survival time as prediction function -- \textbf{Interval-wise feature importance} comparison between estimates obtained with \texttt{DynSurvX}, together with 95\% variability intervals shown as error bars, for CoxSig, DDH, and DySurv (\textbf{three right panels}) at two prediction times $t_{\text{P}} = 2.35$ (\textbf{top}) and $t_{\text{P}} = 3.95$ (\textbf{bottom}).}
    \label{fig:PBC_shap_longi_med}
\end{figure}

\begin{figure}[!h]
    \centering
    \includegraphics[width=\linewidth]{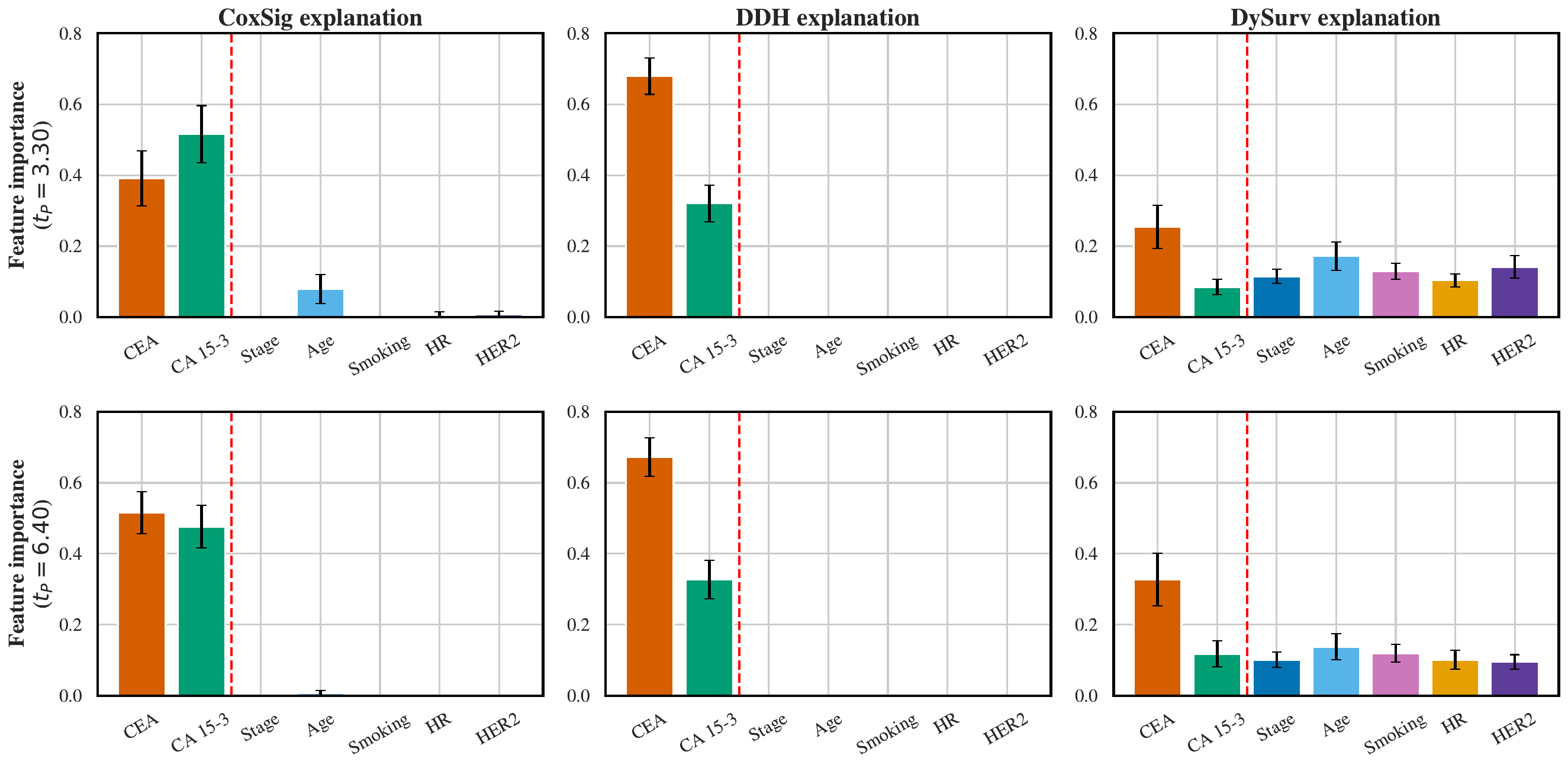}
    \caption{MSK-Chord dataset + median survival time as prediction function -- \textbf{Feature importance} comparison between estimates obtained with \texttt{DynSurvX}, together with 95\% variability intervals shown as error bars, for CoxSig, DDH, and DySurv (\textbf{three right panels}) at two prediction times $t_{\text{P}} = 3.3$ (\textbf{top}) and $t_{\text{P}} = 6.4$ (\textbf{bottom}).}
    \label{fig:MSK_shap_feat_med}
\end{figure}

\begin{figure}[!h]
    \centering  \includegraphics[width=\linewidth]{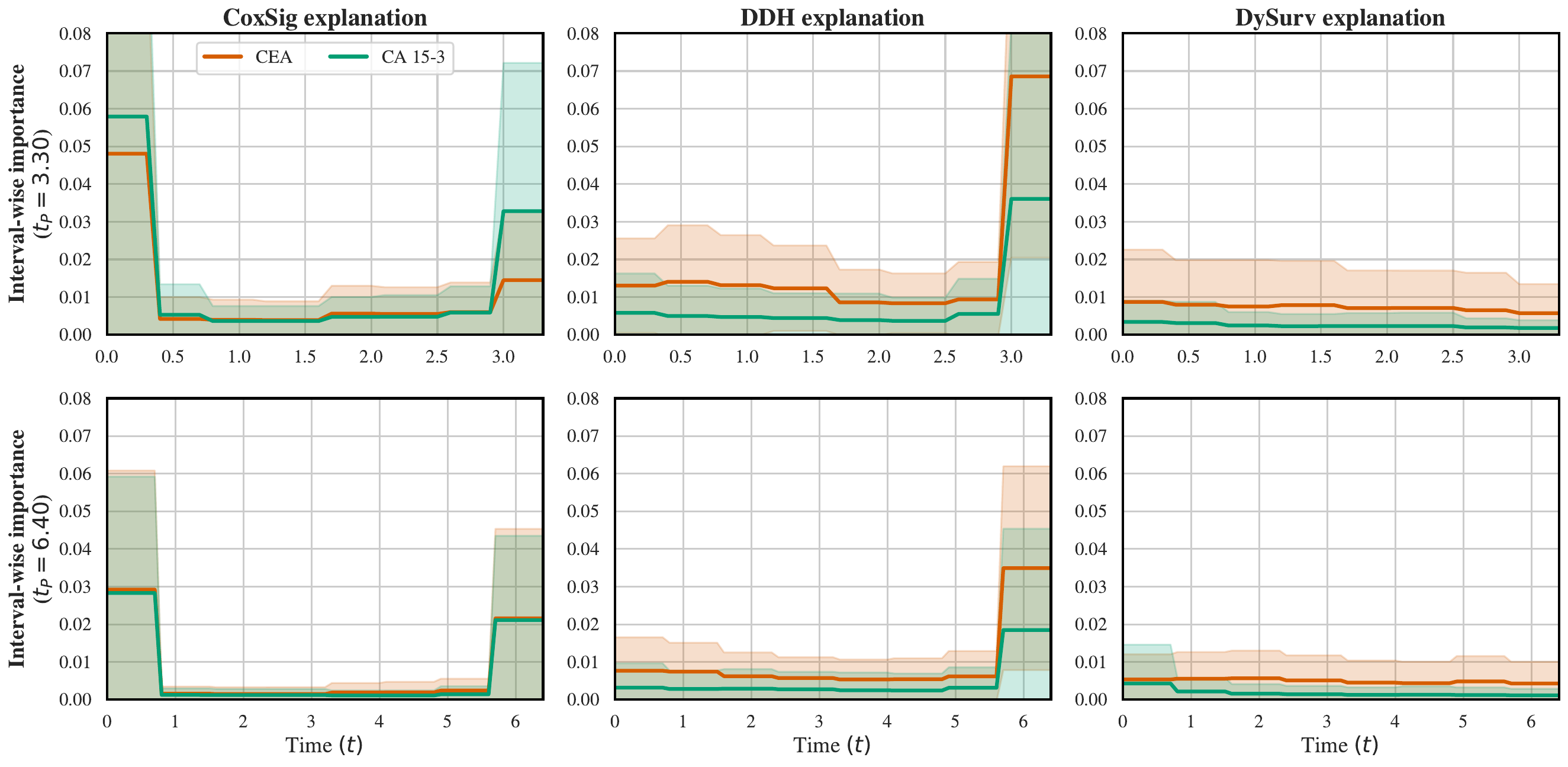}
    \caption{MSK-Chord dataset + median survival time as prediction function -- \textbf{Interval-wise feature importance} comparison between estimates obtained with \texttt{DynSurvX}, together with 95\% variability intervals shown as error bars, for CoxSig, DDH, and DySurv (\textbf{three right panels}) at two prediction times $t_{\text{P}} = 3.3$ (\textbf{top}) and $t_{\text{P}} = 6.4$ (\textbf{bottom}).}
    \label{fig:MSK_shap_longi_med}
\end{figure}

\newpage
\subsection{Sensitive analysis on $\delta_1$}
\label{apd:sensitive_delta1}

The two-phase design may discard longitudinal features with low overall importance but strong localized temporal effects before refinement. To assess robustness, we vary $\delta_1$ around the elbow-selected value.

On PBCseq using DDH, feature- and interval-wise importance patterns remain largely unchanged (Figure~\ref{fig:PBC_DDH_delta1}), indicating that the selected threshold provides stable explanations while preserving clinically relevant temporal effects. However, this does not eliminate the inherent limitation of feature screening. We therefore recommend treating $\delta_1$ as tunable and performing sensitivity analysis, particularly for sparse longitudinal signals.

\begin{figure}[!h]
    \centering
    \includegraphics[width=\linewidth]{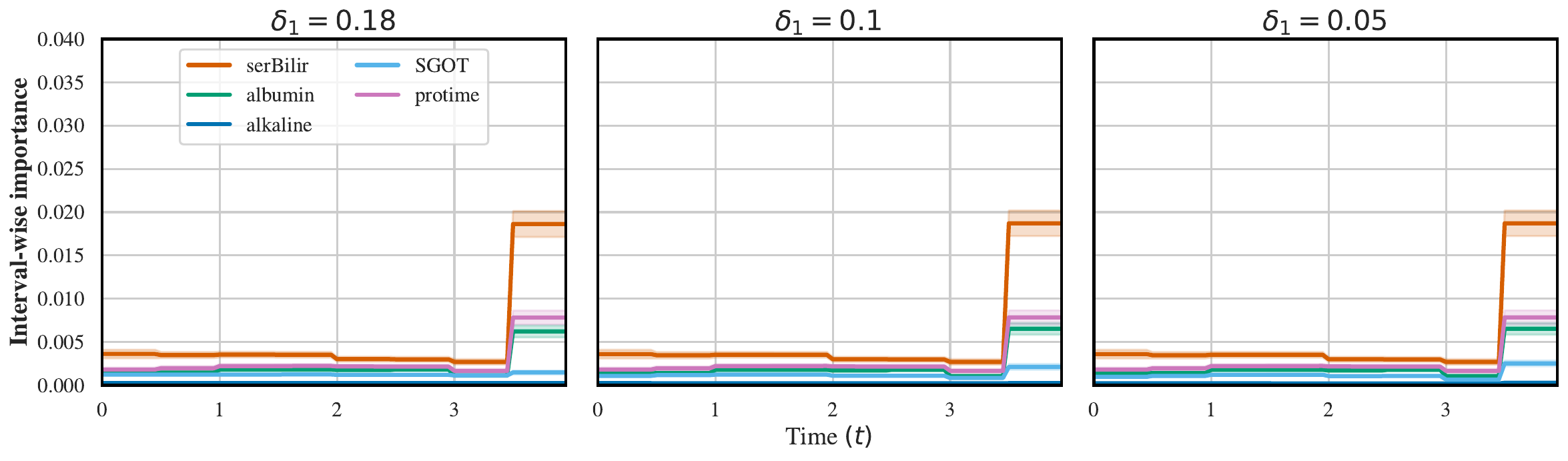}
    \caption{PBCseq dataset -- Sensitivity analysis}
    \label{fig:PBC_DDH_delta1}
\end{figure}

\end{document}